%% file: example_paper_cvpr.tex
\documentclass[10pt,twocolumn,letterpaper]{article}
\makeatletter
\let\@fnsymbol\@arabic
\makeatother

\usepackage{cvpr}              

\usepackage{graphicx}
\usepackage{amsmath}
\usepackage{amssymb}
\usepackage{booktabs}
\RequirePackage{natbib}
\usepackage[pagebackref,breaklinks,colorlinks]{hyperref}

\usepackage{microtype}
\usepackage{dblfloatfix}

\usepackage{hyperref}
\hypersetup{
  colorlinks=false}

\usepackage{amsmath}
\usepackage{amssymb}
\usepackage{mathtools}
\usepackage{amsthm}

\usepackage{amsfonts} 
\usepackage{amsmath}
\usepackage{multirow}
\usepackage{enumitem}
\usepackage{wrapfig}
\usepackage{xcolor}
\usepackage[capitalize,noabbrev]{cleveref}

\theoremstyle{plain}

\theoremstyle{definition}

\theoremstyle{remark}

\newcommand{\shorteqref}[1]{(\ref{#1})}

\usepackage[textsize=tiny]{todonotes}

\newcommand{\rmx}{\mathbf{x}}

\newcommand{\rmz}{\mathbf{z}}
\newcommand{\rmc}{\mathbf{c}}
\newcommand{\rmh}{\mathbf{h}}
\newcommand{\rmm}{\mathbf{m}}
\newcommand{\rmu}{\mathbf{u}}

\definecolor{mygray}{gray}{0.35}
\newcommand{\secondbest}[1]{\color{darkgray}\textbf{#1}}

\usepackage[capitalize]{cleveref}
\crefname{section}{Sec.}{Secs.}
\Crefname{section}{Section}{Sections}
\Crefname{table}{Table}{Tables}
\crefname{table}{Tab.}{Tabs.}

\def\cvprPaperID{*****} 
\def\confName{CVPR}
\def\confYear{2022}

\begin{document}

\title{Multivariate Time Series Forecasting  \\ with Latent Graph Inference}

\vspace{40pt}

\author{
Victor Garcia Satorras \thanks{Work done while at Amazon. Correspondence to: Victor Garcia Satorras, \url{v.garciasatorras@uva.nl}. }\\
University of Amsterdam\\
Amsterdam, The Netherlands \\
\and
Syama Sundar Rangapuram\\
Amazon \\
Berlin, Germany
\and
Tim Januschowski $^\text{1}$ \\
Zalando \\
Berlin, Germany
}
\maketitle

\begin{abstract}
This paper introduces a new approach for Multivariate Time Series forecasting that jointly infers and leverages relations among time series. Its modularity allows it to be integrated with current univariate methods. Our approach allows to trade-off accuracy and computational efficiency gradually via offering on one extreme inference of a potentially fully-connected graph or on another extreme a bipartite graph. In the potentially fully-connected case we consider all pair-wise interactions among time-series which yields the best forecasting accuracy. Conversely, the bipartite case leverages the dependency structure by inter-communicating the $N$ time series through a small set of $K$ auxiliary nodes that we introduce. This reduces the time and memory complexity w.r.t. previous graph inference methods from $O(N^2)$ to $O(NK)$ with a small trade-off in accuracy. We demonstrate the effectiveness of our model in a variety of datasets where both of its variants perform better or very competitively to previous graph inference methods in terms of forecasting accuracy and time efficiency.

\end{abstract}

\input{sections/introduction}

\input{sections/background}
\input{sections/related_work}
\input{sections/method}
\input{sections/experiments}

\input{sections/conclusions}

\bibliography{example_paper_cvpr}
\bibliographystyle{icml2022}

\newpage
\onecolumn
\input{sections/appendix}

\end{document}

%% file: sections/introduction.tex

\section{Introduction}
Time Series Forecasting (TSF) has been widely studied due to its practical significance in a variety of applications such as climate modelling \citep{mudelsee2019trend}, supply chain management in retail \citep{larson2001designing,bose2017probabilistic}, market analysis in finance \citep{andersen2005volatility}, traffic control \citep{li2017diffusion} and medicine \citep{kaushik2020ai} among others. 
In TSF, given a sequence of data points indexed over time, we aim to estimate its future values based on previously observed data. Data is often multivariate, meaning that multiple variables vary over time, each variable may not only depend on its own historical values, but also on other variables' past. Efficiently modelling the dependencies among these variables is still an open problem. \vspace{10pt}

Multivariate Time Series (MTS) methods aim to leverage dependencies between variables in order to improve the forecasting accuracy. 
A natural way to model non-linear dependencies in deep learning is via Graph Neural Networks (GNNs) \citep{bruna2013spectral, defferrard2016convolutional, kipf2016semi}. In fact, GNNs have been successfully applied to MTS forecasting \citep{li2017diffusion, yu2017spatio, chen2020multi}, leveraging the relations among different series. But these methods require a pre-defined adjacency matrix which may only be available in some specific datasets, for example, traffic datasets, where it can be constructed from the spatial structure of a city. More recently, a family of methods that do not require a pre-defined adjacency matrix have been proposed \citep{wu2019graph, franceschi2019learning, wu2020connecting, shang2021discrete}. In this case, a latent graph representation is inferred while forecasting, allowing to operate on a larger variety of datasets. Our work belongs to this category. 

However, inferring all pairwise relations may come at a higher computational cost, as the number of relations scales quadratically $O(N^2) $ w.r.t. the number of nodes/time series. This limits the scalability to large datasets. Additionally, the above mentioned latent graph inference methods perform message passing updates at every time step iteration which is also expensive. 

To overcome these limitations, we propose a new latent graph inference algorithm for MTS forecasting that is more efficient than previous algorithms while achieving better or competitive performance. We cast the latent graph inference as a modular and easy-to-implement extension to current univariate models. The graph is dynamically inferred for each time series input allowing a more flexible representation than a static graph for the whole dataset. Additionally, we can reduce the complexity from $O(N^2)$ (Fully Connected Assumption) to $O(NK)$ (Bipartite Assumption) where $K \ll N$ for a small trade off in performance.

%% file: sections/background.tex

\section{Background}

\subsection{Time Series Forecasting} \label{sec:background_tsf}

In time series forecasting we aim to estimate a future time series $\rmx_{t+1:T}$ given its past $\rmx_{t_0:t}$ where $t_0 \leq t \leq T$ indexes over time, and (optionally) some context information $\rmc$. For the multivariate case we assume the time series is composed of $N$ variates at a time such that $\rmx_{t_0:T} = \{\rmx_{1, t_0:T}, \dots, \rmx_{N, t_0:T}\} \in \mathbb{R}^{N \times T-t_0+1}$. In this section we distinguish two main categories of time series forecasting methods, Global Univariate and Multivariate.

\textbf{Global Univariate methods}: In this case we only use the past of each univariate to predict its future. However, the model weights $\theta_u$ are shared across all univariate time series. More formally:
\begin{equation} \label{eq:univariate}
    \hat{\rmx}_{i,t+1:T} = f_{u}(\rmx_{i,t_0:t}, \rmc_i; \theta_u)
\end{equation}
where $\hat{\rmx}$ denotes the estimated values, $i \in \{1, \dots, N\}$ indexes over multivariates and $f_{_u}(\cdot)$ is the estimator function with learnable parameters $\theta_u$ shared across time series. Conditioning on the past of each univariate may limit the performance of the forecasting algorithm compared to multivariate ones. Despite that, it simplifies the design of $f_\theta$ and already provides reasonable results. A popular example of a global univariate model is DeepAR~\citep{salinas2020deepar}.

\textbf{Multivariate methods}: Multivariate methods condition on all past data (all $N$ variates) and directly predict the multivariate target. More formally:
\begin{equation}
    \hat{\rmx}_{t+1:T} = f_{m}(\rmx_{t_0:t}, \rmc).
\end{equation}
Different variables may be correlated and/or depend on the same con-founders. For example, in retail forecasting, PPE masks and antibacterial soaps jointly increased in demand during the early days of the COVID-19 pandemic. In traffic forecasting, an increase of the outcome traffic flow in a given neighborhood may result in an increase of the income traffic flow on another one. Modelling these dependencies may improve the forecasting accuracy, but it may come at a cost of higher complexity and hence more expensive algorithms, especially when trying to model all pairwise interactions between variates.

\subsection{Graph Neural Networks} \label{sec:background_gnns}
Graph Neural Networks (GNNs) \citep{bruna2013spectral, defferrard2016convolutional, kipf2016semi} operate directly on graph structured data. They have gained a lot of attention in the last years due to their success in a large variety of domains which benefit from modelling interactions between different nodes/entities. In the context of multivariate time series, GNNs can be used to model the interactions between time series. In this work we consider the type of GNN introduced by \citep{gilmer2017neural}. Given a graph $\mathcal{G} = (\mathcal{V}, \mathcal{E})$ with nodes $v_i \in \mathcal{V}$ and edges $e_{ij} \in \mathcal{E}$,  we define a graph convolutional layer as:
\begin{equation} \label{eq:gnn}
\small
    \rmm_{ij} = \phi_e(\rmh_i^l, \rmh_j^l), 
\qquad
\rmh_i^{l+1} = \phi_h(\rmh_i^l, \sum_{j \in \mathcal{N}(i)} \alpha_{ij}\rmm_{i j}) 
\end{equation}
Where $\phi_e$ and $\phi_h$ are the edge and node functions, usually approximated as Multi Layer Perceptrons (MLPs), $\rmh^l_i \in \mathbb{R}^{\text{nf}}$ is the nf-dimensional embedding of a node $v_i$ at layer $l$ and $\rmm_{ij}$ is the edge embedding that propagates information from node $v_j$ to $v_i$. A GNN is constructed by stacking multiple of these Graph Convolutional Layers $\rmh^{l+1} = \mathrm{GCL}[\rmh^{l}, \mathcal{E}]$. Additionally, in~\shorteqref{eq:gnn} we include $\alpha_{ij} \in (0,1)$ which is a scalar value that performs the edge inference or attention over the neighbors similarly to~\cite{velivckovic2017graph}. Following~\citep{satorras2021n}, we choose this value to be computed as the output of a function $\alpha_{ij}=\phi_\alpha(\rmm_{ij})$ where $\phi_\alpha$ is composed of just a linear layer followed by a sigmoid activation function.

%% file: sections/related_work.tex

\section{Related Work}
Time series forecasting has been extensively studied in the past due to its practical significance with a number of recent overview articles available~\citep{petropoulos2020forecasting,overview20,Lim_2021}. Traditionally, most classical methods are univariate in nature (see e.g.,~\cite{hyndman2017forecasting} for an overview). While some of these have multi-variate extensions (e.g., ARMA and VARMA models), they are limited by the amount of related time series information they can incorporate. Dynamic factor models \citep{geweke1977dynamic, wang2019deepfactors} are fore-runners of a family of models that has recently received more attention, the so-called global models \citep{januschowski19,montero2020principles}, see Section~\ref{sec:background_tsf}. These global models estimate their parameters over an entire panel of time series, 
but still produce a univariate forecast. Many such global models have been proposed building on neural network architectures like RNNs~\citep{salinas2020deepar,Liberty2020,kasun19}, CNNs~\citep{wen2017multi,chen2019probabilistic}, Transformers~\citep{li2019neurips,lim2021temporal,eisenach2020mqtransformer} and also combining classical probabilistic models with deep learning \citep{rangapuram2018deep,kurle2020deep,de2020normalizing}.  However, unlike our method, these global models do not explicitly model the relationship between the time series in the panel.

Most recently, global multi-variate forecasting models have received attention, in particular models that attempt to capture the relationship of the time series via a multi-variate likelihood \citep{rasul2020multi,rasul2021autoregressive,de2020normalizing,multivariate19}. A complementary approach consists of capturing the multi-variate nature of many modern forecasting problems primarily by using a multi-variate time series as input. For this, a natural way to model and exploit the relationship between time series is via GNNs~\citep{bruna2013spectral, defferrard2016convolutional, kipf2016semi}. Even in those scenarios where the adjacency of the graph is explicitly provided, attention or a latent graph can be inferred from the node embeddings such that GNNs can still leverage the structure of the data. Some examples of latent graph inference or attention are \citep{wang2019dynamic} in point clouds, \citep{franceschi2019learning} in semi-supervised graph classification,  \citep{ying2018hierarchical} in hierarchical graph representation learning, \citep{kipf2018neural} in modelling dynamical systems, \citep{kazi2020differentiable} in zero-shot learning and 3D point cloud segmentation, \citep{garcia2017few, kossen2021self} in image classification, \citep{cranmer2020discovering} in inferring symbolic representations and \citep{fuchs2020se, satorras2021n} in molecular property prediction.

In MTS forecasting, \citet{lai2018modeling,shih2019temporal} provide some of the first deep learning approaches designed to leverage pair-wise dependencies among time series. More recent methods~\citep{li2017diffusion,yu2017spatio,seo2018structured,zhao2019t} are built in the intersection of GNNs and time series forecasting but they require a pre-defined adjacency matrix. Lately, new methods that infer a latent graph from the node embeddings have been introduced in MTS forecasting \citep{wu2020connecting, shang2021discrete, cao2021spectral}, and in MTS anomaly detection \citep{zhang2020correlation, deng2021graph}. These methods can be applied to any dataset even when there is not an explicitly defined adjacency matrix. The main limitation is the computational cost, which occurs in the edge inference and scales quadratically $O(N^2)$  w.r.t the number of nodes/time series $N$ or $O(N^3)$ in \citep{wu2020connecting, cao2021spectral}.

Our approach is related to \citep{wu2020connecting, shang2021discrete}, but in contrast (i) our latent graph is dynamically inferred for each input instead of static for the whole dataset; (ii) it only requires the message exchange at time step $t$ which renders the graph operation cheaper and modular; and (iii) it optionally reduces the number of edges from $O(N^2)$ to $O(NK)$ when using a bipartite assumption ($K \ll N$).

\paragraph{Linear vs Non-linear message passing} 
A shared potential limitation among the mentioned methods that use GNNs for MTS Forecasting \citep{yu2017spatio, li2017diffusion, seo2018structured, zhao2019t, wu2019graph, wu2020connecting, shang2021discrete, cao2021spectral} is the use of a linear function in the edge operation \citep{defferrard2016convolutional, kipf2016semi}. In contrast, we allow for non-linearities in our model.  Table~\ref{tab:gnns_comparison} provides an overview comparing these two alternatives by writing the linear graph convolutional layer in the same format as the non-linear one \citep{gilmer2017neural} from Section \ref{sec:background_gnns}. Where $\theta_e,\theta_h \in \mathbb{R}^{\text{nf} \times \text{nf}}$ are learnable parameters.

\vspace{-4pt}
\begin{table}[h!]
\footnotesize
\begin{center} 
\begin{tabular}{ c  c c } 
\toprule
  & Non-linear & Linear\\
  \midrule
  Edge & $\rmm_{ij} = \phi_e(\rmh_i^l, \rmh_j^l)$ & $\rmm_{ij} = \mathbf{\theta}_e^\intercal \rmh_j $ \\
  \midrule
Aggr &  \multicolumn{2}{c }{$\rmm_i = \sum_{j \in \mathcal{N}(i)} \rmm_{i j}$} \\

  \midrule
  Node & $\rmh_i^{l+1} = \phi_h(\rmh_i^l, \rmm_i) $ & $\rmh_i^{l+1}  =  \sigma(\rmm_i + \theta_h^\intercal\rmh_h^l) $  \\

    \bottomrule
\end{tabular}
\vspace{-3pt}
\caption{Non-linear vs Linear Graph Conv. Layer comparison.}
\label{tab:gnns_comparison} 
\end{center}
\vspace{-14pt}
\end{table}

The linear graph convolutional layer is commonly written in matrix form as we show in Appendix~\ref{ap:linear_graph}. Note that while the non-linear alternative approximates the edge operation with an MLP $\phi_e$, in the linear case the message operation among two nodes ($i,j$) only depends on node $\rmh_j$ and it is agnostic to the identity of the receiver node $i$. This can be problematic when inferring relations among two nodes since the message operation only depends on one of them. Therefore, when no adjacency matrix is provided, an attention mechanism that depends on both $i,j$ can address some of these limitations $\rmm_{ij}= \alpha(i,j) \theta_e^\intercal \mathbf{h}_j$. However, note that in the non-linear case, $\phi_e$ is a universal approximator that can be interpreted as a generalization of the linear case with attention ${\phi}_e(\rmh_i, \rmh_j) = \alpha(i,j) \theta_e^\intercal \mathbf{h}_j$. In practice, even if the non-linear case can more flexibly model $\rmm_{ij}$, it has been shown in previous literature that attention or edge inference \citep{vaswani2017attention, satorras2021n} can be a strong inductive bias which in our case is able to infer the graph structure as we will show in the experiments section.


%% file: sections/method.tex

\section{Method} \label{sec:method}

\begin{figure*}[t]
\center
\includegraphics[width=0.9\linewidth]{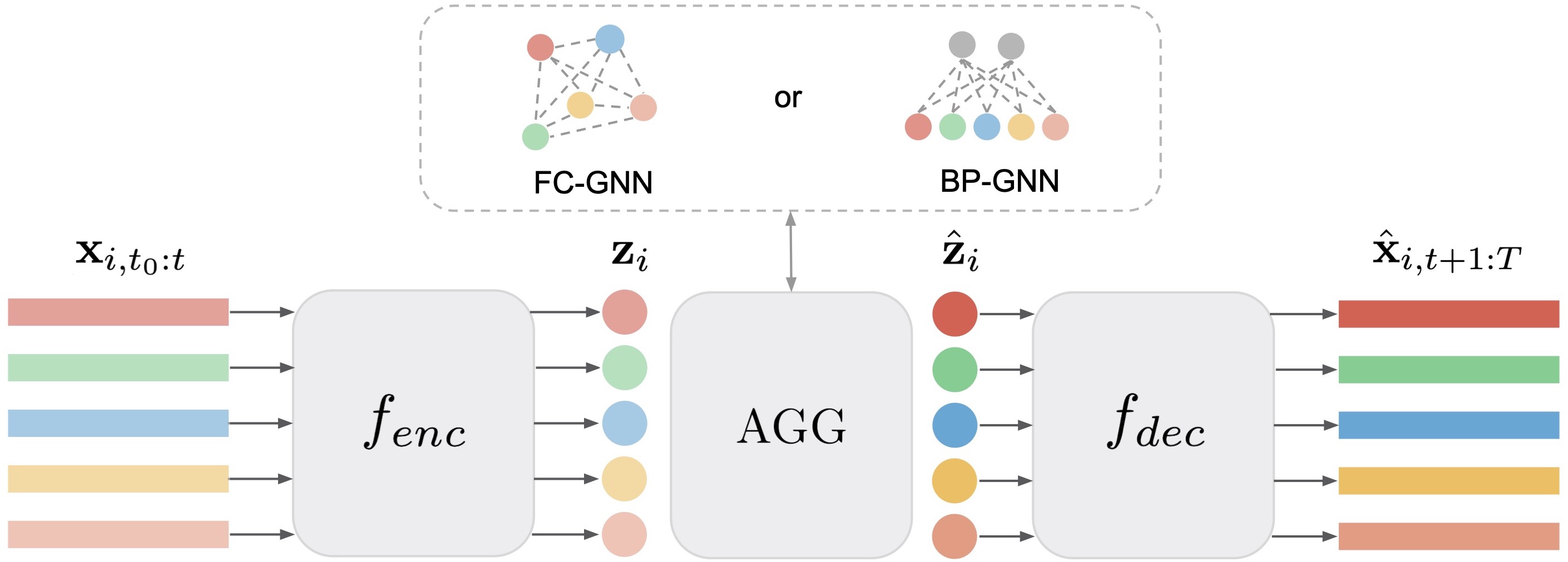}
\caption{Illustration of the presented method under fully connected and bipartite graph assumptions.}
\label{fig:main_model}
\end{figure*}


In Section \ref{sec:background_tsf} we discussed two families of forecasting methods: Global Univariate Models and Multivariate Models. We cast our multivariate algorithm as a modular extension of the univariate case from ~\shorteqref{eq:univariate}. We can break down the Univariate Model in two main steps $f_u = f_{enc} \circ f_{dec}$ such that $\rmx_{i,t_0:t} \stackrel{f_\text{enc }}{\longrightarrow} \rmz_i \stackrel{f_\text{dec }}{\longrightarrow} \hat{\rmx}_{i,t+1:T}$, where $f_{\text{enc}}$ encodes the input signal $\rmx_{i, t_0, t}$ (and optionally some context information $\rmc_i$) into an embedding $\rmz_i$, and $f_\text{dec }$ estimates the future signal from this embedding. In our method, we include a multivariate aggregation module $\mathrm{AGG}$ in between $f_{enc}$ and $f_{dec}$ that propagates information among nodes in the latent space $\rmz = \{\rmz_1, \dots, \rmz_N \}$. This aggregation module takes as input the embedding $\rmz_{i}=f_{enc}( \rmx_{i,t_0:t}, \rmc_i)$, and outputs a modified embedding $\hat{\rmz} = \mathrm{AGG}(\rmz)$ where information has been propagated among nodes. These new embeddings are then passed as input to the decoder $f_{dec}$. The resulting algorithm is:
\begin{align}
    \text{Univariate Encoder}\quad & \rmz_{i}=f_{enc}( \rmx_{i,t_0:t}, \rmc_i) \label{eq:model_encoder}\\
   \text{Multivariate extension}\quad & \hat{\rmz} = \mathrm{AGG}(\rmz) \label{eq:model_aggr} \\
   \text{Univariate Decoder}\quad & \hat{\rmx}_{i, t+1:T} = f_{dec}(\hat{\rmz}_i) \label{eq:model_dec}
\end{align}
Notice the overall model is multivariate but $f_{enc}$ and $f_{dec}$ remain univariate, allowing a modular extension from current univariate methods to multivariate. Additionally, in contrast to the most recent methods \citep{wu2020connecting, shang2021discrete}, our model does not propagate information among nodes at every time step $[t_0,\dots, t]$ but only in the $\mathrm{AGG}$ module. This makes the algorithm cheaper since the message propagation step is usually an expensive operation. Additionally, we experimented including a unique identifier of the time series in the encoded signal $\rmz_i$ as context information $\rmc_i = \text{id}$, which resulted in a significant improvement in accuracy as we will show in the experiments section. In the following subsections we propose two different GNN configurations for the $\mathrm{AGG}$ module depending on the assumed structure of the inferred graph. One is a potentially fully connected graph and the other is a bipartite graph that communicates the N time series through a small set of $K$ auxiliary nodes that we introduce, each has its benefits and weaknesses regarding performance and scalability.  An illustration of the whole algorithm is presented in Figure \ref{fig:main_model}.

\textbf{Fully connected graph assumption | FC-GNN}

From the two configurations, this is the most straightforward to implement given a standard GNN. It results in good performance, but its complexity scales $O(N^2)$ with respect to the number of nodes $N$. Despite this, it still resulted in faster computation times than previous $O(N^2)$ methods.
We can start by defining a fully connected graph $\mathcal{G} = \{\mathcal{V}, \mathcal{E}\}$ where all nodes exchange messages among each other such that $e_{ij}=1$ for all $e_{ij} \in \mathcal{E}$. Each time series embedding $\rmz_i$ obtained from the Univariate Encoder~\shorteqref{eq:model_encoder} is associated with a node of the graph $v_i \in \mathcal{V}$. Then, we can directly use the GNN defined in Section~\shorteqref{sec:background_gnns} as the aggregation module $\mathrm{AGG}$ where each embedding $\rmz_i$ is provided as the input $\rmh_i^0$ to to the GNN~\shorteqref{eq:gnn}, after the GNN runs for $L$ layers, the output node embedding $\rmh^L_i$ is provided as the input $\hat{\rmz}_i$ to the Univariate Decoder~\shorteqref{eq:model_dec}. Moreover, despite the fully connected assumption, the GNN infers attention weights $\alpha_{ij} \in (0, 1)$ \shorteqref{eq:gnn} for each edge and input sample that "gate" the exchanged messages $\rmm_i = \sum_{i \neq j}\alpha_{ij}\rmm_{ij}$. This can be interpreted as dynamically inferring the graph. The reason is that in GNNs we can write the message aggregation as $\rmm_i = \sum_{j \in \mathcal{N}(i)} \rmm_{ij}=\sum_{j  \neq j} e_{ij}\rmm_{ij}$ where $e_{ij}$ is 1 if the edge exists or 0 otherwise. We can see that the following expressions $\sum_{i \neq j}e_{ij}\rmm_{ij} \approx  \sum_{i \neq j}\alpha_{ij}\rmm_{ij}$ becomes equivalent when the soft estimation $\alpha_{ij}=\phi_\alpha(\rmm_{ij})$ approximates/infers the underlying graph $e_{ij}$ structure from the data. In the experiments section, we name this model a Fully Connected Graph Neural Network (FC-GNN). 

\textbf{Bipartite graph assumption | BP-GNN}

In the previous section we presented a GNN method that exchanges information among time series under a fully connected graph assumption which computationally scales $O(N^2)$. In this section we introduce a bipartite graph assumption (BP-GNN) which reduces the complexity to $O(N K)$, $K$ being a parameter of choice $K \ll N$. To accomplish this, we define a bipartite graph $\mathcal{G} = (\mathcal{Y}, \mathcal{U}, \mathcal{E})$, where $\mathcal{Y}$ is a set of $N$ nodes corresponding to the $N$ time series and $\mathcal{U}$ is a set of $K$ auxiliary nodes. Nodes $\mathcal{Y}$ have associated the time series embeddings $\rmz = \{\rmz_1, \dots \rmz_N\}$ and auxiliary nodes $\mathcal{U}$ have associated the embeddings $\rmu = \{\rmu_1, \dots \rmu_K\}$ which are free learnable parameters initialized as gaussian. Edges $\mathcal{E}$ interconnect all nodes between the two subsets $\{\mathcal{Y}$, $\mathcal{U}\}$, but there are no connections among nodes within the same subset. This results in $2NK$ edges.

The message passing scheme works in the following way. We input into the GNN the union of the two node subsets $\mathcal{V} = \mathcal{Y} \cup \mathcal{U}$. Specifically, the input embedding $\rmh^0$ defined in Equation \ref{eq:gnn} is the concatenation of the  time series embeddings $\rmz$ with the auxiliary node embeddings $\rmu$ (i.e. $\rmh^0 = \rmz || \rmu$). Then, messages follow an asynchronous schedule, first information is propagated from the times series nodes to the auxiliary nodes $\mathcal{Y} \xrightarrow{} \mathcal{U}$, next the other way around $\mathcal{U} \xrightarrow{} \mathcal{Y}$.

\begin{table}[h] \small
\setlength{\tabcolsep}{3pt}
\begin{center} 
\begin{tabular}{c|c}
\toprule
Step 1 & Step 2\\
$\mathcal{Y} \rightarrow \mathcal{U}$ & $\mathcal{U} \rightarrow \mathcal{Y}$ \\
\midrule 
$i \in \mathcal{U} \quad , \quad j \in \mathcal{Y}$ & $i \in \mathcal{Y} \quad , \quad j \in \mathcal{U}$\\
\midrule
\begin{math}
\begin{aligned}
\rmm_{ij} &= \phi_{e_1}(\rmh_i^l, \rmh_j^l, a_{ij}) \\
    \rmm_i &= {\textstyle \sum_{j \in \mathcal{Y}} \phi_{\alpha_1}(\rmm_{i j})\rmm_{i j}} \\
    \rmh_i^{l+1} &= \phi_{h_1}(\rmh_i^l, \rmm_i) \\
\end{aligned}
\end{math}
 &
 \begin{math}
\begin{aligned}
\rmm_{ij} &= \phi_{e_2}(\rmh_i^l, \rmh_j^l, a_{ij}) \\
    \rmm_i &= {\textstyle \sum_{j \in \mathcal{U}}} \phi_{\alpha_2}(\rmm_{i j})\rmm_{i j} \\
    \rmh_i^{l+1} &= \phi_{h_2}(\rmh_i^l, \rmm_i) \\
\end{aligned}
\end{math}\\
\bottomrule
\end{tabular}
\caption{BP-GNN formulation.}
\label{tab:eq_bp} 
\end{center}
\vspace{-12pt}
\end{table}

We have conceptually defined the Bipartite Graph Neural Network. In Table \ref{tab:eq_bp} we introduce the equations that formally define it as an extension of the standard GNN Equation
\ref{eq:gnn}. Notice that it can be simply formulated as a two steps process where the indexes $i,j$ belong to each one of the subsets $\mathcal{U}$ or $\mathcal{Y}$ depending on the direction of the messages ($\mathcal{Y} \rightarrow \mathcal{U}$ or $\mathcal{U} \rightarrow \mathcal{Y}$). Additionally, we used different learnable parameters between Step 1 and Step 2 in the modules $\phi_e$, $\phi_{\alpha}$ and $\phi_h$ since it resulted in better performance than sharing parameters. Following, we define the adjacency matrices corresponding to the two message passing steps (assuming all inference parameters $\alpha_{ij} = 1$):

\begin{equation}
\small
  A_1 =  \begin{vmatrix} 
    0_{N \times N} & 0_{N \times K}   \\
   1_{K \times N} & 0_{K \times K} \\
   \end{vmatrix} 
,\quad
  A_2 =  \begin{vmatrix} 
  0_{N \times N} & 1_{N \times K}   \\
   0_{K \times N} & 0_{K \times K} \\
   \end{vmatrix}
       \label{eq:bp_adjacencies}
\end{equation}

$A_1$ refers to $\mathcal{Y} \xrightarrow{} \mathcal{U}$ and $A_2$ refers to $\mathcal{U} \xrightarrow{} \mathcal{Y}$. The product of these two matrices $\tilde{A} = A_2A_1$ (Appendix \ref{eq:bp_adjacencies}) defines the sum of all paths that communicate the time series nodes $\mathcal{Y}$ among each other through the auxiliary nodes $\mathcal{U}$. Notice that for $K>0$ all nodes $\mathcal{Y}$ can potentially communicate among them being more efficient than FC-GNN as long as $0  \leq K < N/2$.

\textbf{Architecture details}

Recall that our method is composed of three main modules, the encoder $f_{enc}$, the decoder $f_{dec}$ and the aggregation module $\mathrm{AGG}$. We choose to use relatively simple networks as encoder and decoder. The decoder $f_{dec}$ is defined as a Multi Layer Perceptron (MLP) with a single hidden layer and a residual connection in all experiments. The encoder $f_{enc}$ is also defined as an MLP for METR-LA, PEMS-BAY and our synthetic datasets and as a Convolutional Neural Network (CNN) for the other datasets since these require a larger encoding length and the translation equivariance of CNNs showed to be more beneficial. The encoder $f_{enc}$, first encodes the input signal $\rmx_{i, t_0:t}$ to an embedding vector by using the mentioned MLP or CNN, and then concatenates a unique identifier ($\rmc_i = \text{id}$) to the obtained embedding vector resulting in $\rmz_i$. $\rmc_i$ could (optionally) include additional context information if it was provided in the dataset. The combination of these simple networks with our proposed aggregation module $\mathrm{AGG}$ fully defines our model. 

The aggregation module $\mathrm{AGG}$ was defined under two different assumptions, the Fully Connected Graph assumption (FC-GNN) and the Bipartite Graph assumption (BG-GNN). In both cases the GNN is fully parametrized by the networks 
$\phi_e$, $\phi_h$ and $\phi_\alpha$. $\phi_e$ consists of a two layers MLP, $\phi_h$ is a one layer MLP with a skip connection from he input to the output and $\phi_\alpha$ is just a linear layer followed by a Sigmoid activation function.
 All architecture choices are explained in more detail in Appendix~\ref{ap:architecture_choices}. 
We optimized the Mean Absolute Error as the loss $\mathcal{L}=l(\hat{\rmx}_{ t+1:T}, \rmx_{t+1:T})$ for training.



%% file: sections/experiments.tex

\section{Experiments} \label{sec:experiments}
\subsection{Datasets and Baselines}

\begin{table}[h!]
\vspace{-5pt}
\scriptsize
\begin{center} 
\begin{tabular}{l | cc cc  }
\toprule
 & \#Nodes & \# Samples & Context length & Pred. length\\
 \midrule
 METR-LA & 207 & 34,272 & 12 & 12 \\
PEMS-BAY & 325 & 52,116 & 12 & 12 \\
\midrule
Solar-Energy & 137 & 52,560 & 168 & 1\\
Traffic & 862 & 17,544 & 168 & 1\\
Electricity & 321 & 26,304 & 168 & 1\\
Exchange-Rate & 8  &  7,588 & 168 & 1\\
    \bottomrule
\end{tabular}
\vspace{-5pt}
\caption{Dataset specifications.}
\label{tab:mtgnn_results} 
\label{table:datasets}
\end{center}
\vspace{-15pt}
\end{table}

We first evaluate our method on METR-LA and PEMS-BAY datasets~\citep{li2017diffusion} which record traffic speed statistics on the highways of Los Angeles county and the Bay Area respectively. 
We also consider the publicly available Solar-Energy, Traffic, Electricity and Exchange-Rate data sets. Specifications for each dataset are presented in Table \ref{table:datasets} and further detailed in Appendices \ref{ap:metr_pems} and \ref{ap:single_step_impl_details}.
We compare to the following three main types of baselines:

\begin{table*}[t]
\scriptsize
\setlength{\tabcolsep}{3.5pt}
\begin{center}
\begin{tabular}{l | ccc | ccc | ccc | ccc | ccc | ccc  }
\toprule
 & \multicolumn{9}{c|}{METR-LA} & \multicolumn{9}{c}{PEMS-BAY}  \\
 & \multicolumn{3}{c}{15 min} & \multicolumn{3}{c}{30 min} & \multicolumn{3}{c|}{60 min}& \multicolumn{3}{c}{15 min} & \multicolumn{3}{c}{30 min} & \multicolumn{3}{c}{60 min} \\
 \midrule
  & \tiny MAE & \tiny RMSE & \tiny MAPE & \tiny MAE & \tiny RMSE & \tiny MAPE & \tiny MAE & \tiny RMSE & \tiny MAPE & \tiny  MAE & \tiny RMSE & \tiny  \tiny MAPE & \tiny MAE & \tiny RMSE & \tiny MAPE & \tiny MAE & \tiny RMSE & \tiny MAPE \\
 \midrule
x5 Linear / AR & 3.81 & 8.80 & 9.13\% & 4.94 & 11.14 & 12.17\% & 6.30 & 12.91 & 16.72\% & 1.59 & 3.41 & 3.27\% &  2.15 & 4.87 & 4.77 \% & 2.97 & 6.65 & 7.03\% \\
 \midrule

 DCRNN & 2.77 & 5.38 & 7.30\% & 3.15 & 6.45 & 8.80\% & 3.60 & 7.60 & 10.50\% & 1.38 & 2.95 & 2.90\% & 1.74 & 3.97 & 3.90\% & 2.07 & 4.74 & 4.90\% \\
 STGCN & 2.88 & 5.74 & 7.6\% & 3.47 & 7.24 & 9.6\% & 4.59 & 9.40 & 12.7\% & 1.36 & 2.96 & 2.9\% & 1.81 & 4.27 & 4.2\% & 2.49 & 5.69 & 5.8\%\\
 MRA-BGCN & 2.67 & \textbf{5.12} & \textbf{6.8}\% & 3.06 & 6.17 & 8.3\% & 3.49 & 7.30 & 10.0\% & \textbf{1.29} & \textbf{2.72} & 2.9\% & \textbf{1.61} & \textbf{3.67} & 3.8\% & \textbf{1.91} & \textbf{4.46} & 4.6\% \\
 \midrule
   Graph WaveNet* & 2.69 & 5.15 & 6.90\% & 3.07 & 6.22 & 8.37\% & 3.53 & 7.37 & 10.01\% & 1.30 & 2.74 & \textbf{2.73\%} & 1.63 & 3.70 & \textbf{3.67\%} & 1.95 & 4.52 & 4.63\% \\
 LDS & 2.75 & 5.35 & 7.1\% & 3.14 & 6.45 & 8.6\% & 3.63 & 7.67 & 10.34\% & 1.33 & 2.81 & 2.8\% & 1.67 & 3.80 & 3.8\% & 1.99 & 4.59 & 4.8 \% \\
MTGNN & 2.69 & 5.18 & 6.86\% & 3.05 & 6.17 & 8.19\% & 3.49 & 7.23 & 9.87\% & 1.32 & 2.79 & 2.77\% & 1.65 & 3.74 & 3.69\% & 1.94 & 4.49 & \textbf{4.53}\%
  \\
        GTS & 2.64 & 5.19 & 6.79\% & 3.06 & 6.30 & 8.24\% & 3.56 & 7.55 & 9.95\% & 1.35& 2.84 & 2.85\% & 1.67 & 3.82 & 3.80\% & 1.96 & 4.53 & 4.62\% \\
     \midrule
       \textit{Ablation study} &    \\
 NE-GNN \textit{(w/o id)} & 2.80 & 5.73 & 7.50\% & 3.40 & 7.15 & 9.74\% & 4.22 & 8.79 & 13.06\% & 1.40 & 3.03 & 2.92\% & 1.85 & 4.25 & 4.21\% & 2.39 & 5.50 & 5.93\%  \\
 FC-GNN \textit{(w/o id)} & 2.77 & 5.65 & 7.39\% & 3.36 & 7.02 & 9.59\% & 4.14 & 8.64 & 12.70\% & 1.39 & 3.00 & 2.88\% & 1.82 & 4.18 & 4.11\% & 2.32 & 5.35 & 5.71\%  \\
 NE-GNN & 2.69 & 5.57 & 7.21\% & 3.14 & 6.74 & 9.01\% & 3.62 & 7.88 & 10.94\% & 1.36 & 2.88 & 2.86\% & 1.72 & 3.91 & 3.93 \% & 2.07 & 4.79 & 5.04\% \\
 \midrule
 \textit{Our models} & \\
 FC-GNN & \textbf{2.60} & 5.19 & \textbf{6.78\%} & \textbf{2.95} & \textbf{6.15} & \textbf{8.14\%} & \textbf{3.35} & \textbf{7.14} & \textbf{9.73\%} & 1.33 & 2.82 & 2.79\% & 1.65 & 3.75 & \textbf{3.72\%} & 1.93 & \textbf{4.46} & \textbf{4.53}\%    \\
  BP-GNN (K=4) & 2.64 & 5.37 & 7.07\% & 3.02 & 6.42 & 8.46\% &  3.40 & 7.32 & 9.91\% & 1.33 &  2.82 & 2.80\% & 1.66 & 3.78 & 3.75\% & 1.94 & \textbf{4.46} & 4.57\% \\
    \bottomrule
\end{tabular}
\caption{Benchmark on METR-LA and PEMS-BAY datasets. Mean Absolute Error (MAE), Root Mean Squared Error (RMSE) and Mean Absolute Percentage Error (MAPE) are reported for different time horizons \{15, 30, 60\} minutes. Results have been averaged over 5 runs.}
\vspace{-10pt}
\label{tab:overview_results}
\end{center}
\end{table*}

\begin{itemize}[leftmargin=*]
    \item Univariate: In this case a single model is trained for all time series but they are treated independently without message exchange among them. These baselines include a simple linear Auto Regressive model (AR) and a variation of our FC-GNN method that we called NE-GNN where all edges have been removed such that the aggregation function $\mathrm{AGG}$ becomes equivalent to a multilayer perceptron defined by $\phi_h$~\shorteqref{eq:gnn}. 
    \item Multivariate with a known graph: These methods require a previously defined graph, therefore they are restricted to those datasets where an adjacency matrix can be pre-defined (e.g. METR-LA, PEMS-BAY). From this group we compare to DCRNN \citep{li2017diffusion}, STGCN \citep{yu2017spatio} and MRA-BGCN \citep{chen2020multi}.
    
    \item Multivariate with graph inference or attention: These methods exchange information among different time series by simultaneously inferring relations among them or by attention mechanisms. From this group we compare to LDS \citep{franceschi2019learning},
    LST-Skip \citep{lai2018modeling}, TPA-LSTM \citep{shih2019temporal}, MTGNN \citep{wu2020connecting} and GTS \citep{shang2021discrete}. Graph WaveNet \citep{wu2019graph} also belongs to this group but unlike the others it jointly uses a pre-defined adjacency matrix. Comparisons to NRI \citep{kipf2018neural} can be found in previous literature \citep{shang2021discrete, zugner2021study}. We additionally include a variation of our FC-GNN without unique node ids, we denote it by \textit{(w/o id)} beside the model name. GTS numbers have been obtained by averaging over 3 runs its official implementation.
\end{itemize}

\subsection{Main results} \label{sec:main_results}

In this section we evaluate our method in METR-LA and PEMS-BAY datasets. For this experiment we used the training setup from GTS \citep{shang2021discrete} which uses the dataset partitions and evaluation metrics originally proposed in \citep{li2017diffusion}. All our models have been trained by minimizing the Mean Absolute Error (MAE) between the predicted and ground truth samples. The reported metrics are MAE, Root Mean Squared Error (RMSE) and Mean Absolute Percentage Error (MAPE) from \citep{li2017diffusion}. All metrics have been averaged over 5 runs. All our models (FC-GNN, BP-GNN and NE-GNN) contain 2 graph convolutional layers, 64 features in the hidden layers, Swish activation functions \citep{ramachandran2017searching} and have been trained with batch size 16. The number of auxiliary nodes in BP-GNN was set to $K=4$.  Time experiments report the average forward pass in seconds for a batch size 16 in a Tesla V100-SXM GPU. Further implementation details are provided in Appendix \ref{ap:impl_metr_pems}.

\textbf{Results} are reported in Tables \ref{tab:overview_results} and \ref{tab:time_results}, standard deviations in Appendix \ref{ap:metr_pems_std}.
FC-GNN outperforms other methods in most metrics while being computationally cheaper than previous works. On the other hand, BP-GNN performs very competitively w.r.t. previous works (even outperforming all previous methods in some metrics) but with a vast decrease in computation. Furthermore, these performances are achieved without explicitly providing the structural information of the city (i.e. pre-defined adjacency) unlike in those methods that require it (e.g. MRA-BGCN) or in Graph Wavenet that optionally uses it. Additionally, note the performance gap between FC-GNN and NE-GNN is larger when including a unique identifier of the nodes. This means the network can better leverage the information exchange among nodes when they are uniquely identified, but it still benefits from the message passing scheme when they are not uniquely identified (w/o id) thanks to the dynamical inference.
 \begin{table}[h!]
\vspace{-5pt}
\footnotesize
\renewcommand{\arraystretch}{0.9}
\begin{center}
\begin{tabular}{l | cc  }
\toprule
 & \multicolumn{2}{c}{Forward Time (s)} \\
 & METR-LA & PEMS-BAY \\
 \midrule
 Linear & .0002 & .0002\\
 \midrule
 DCRNN & .2559  & .2754 \\
 \midrule
 Graph WaveNet & .0500 & .0673\\
 MTGNN & .0160 & .0371\\
 GTS & .0869 & .1087\\
 \midrule
 NE-GNN & .0033 & .0047\\
 FC-GNN & .0108 & .0253\\
 BP-GNN (K=4) & .0044 & .0046\\
    \bottomrule
\end{tabular}
\vspace{-6pt}
\caption{Forward time in seconds for different methods.}
\label{tab:time_results}
\end{center}
\vspace{-5pt}
\end{table}
Time results are reported in Table \ref{tab:time_results}. We include all methods from the previous Table \ref{tab:overview_results} that have a publicly available implementation in METR-LA and PEMS-BAY datasets. BP-GNN is the most efficient algorithm in both METR-LA and PEMS-BAY by a large margin. FC-GNN is also more efficient than previous methods in both datasets but it is still limited by the $O(N^2)$ scalability. In the next Section \ref{sec:single_step} we will see that BP-GNN becomes even more efficient for larger graphs compared to the other methods thanks to its better scalability. 



\begin{table*}[t]
\scriptsize
\setlength{\tabcolsep}{4pt}

\begin{center}
\begin{tabular}{ll | c | cccc | cccc | cccc | cccc}
\toprule
\multicolumn{2}{c|}{Dataset} &
&\multicolumn{4}{|c|}{Solar-Energy} & \multicolumn{4}{|c|}{Traffic} &
\multicolumn{4}{|c|}{Electricity} &
\multicolumn{4}{|c}{Exchange-Rate} \\
\midrule
\multicolumn{2}{c|}{} &
\multicolumn{1}{c|}{}&
\multicolumn{4}{c}{Horizon} & \multicolumn{4}{|c|}{Horizon} & \multicolumn{4}{|c|}{Horizon} & \multicolumn{4}{|c}{Horizon} \\
Methods & Metrics & \#Top2 &
3 & 6 & 12 & 24 &
3 & 6 & 12 & 24 &
3 & 6 & 12 & 24 &
3 & 6 & 12 & 24\\
\midrule
\multirow{2}{*}{AR} & RSE & (0) & .2435 & .3790 & .5911 & .8699 & .5991 & .6218 & .6252 & .6293 & .0995 & .1035 & .1050 & .1054 & .0228 & .0279 & .0353 & .0445 \\
& CORR & (0) & .9710 & .9263 & .8107 & .5314 & .7752 & .7568 & .7544 & .7519 & .8845 & .8632 & .8591 & .8595 & .9734 & .9656 & .9526 & .9357 \\
\multirow{2}{*}{RNN-GRU} & RSE & (0) & .1843 & .2559 & .3254 & .4643 & .4777 & .4893 & .4950 & .4973 & .0864 & .0931 & .1007 & .1007 & .0226 & .0280 & .0356 & .0449 \\
& CORR & (0)  & .9843 & .9690 & .9467 & .8870 & .8721 & .8690 & .8614 & .8588 & .9283 & .9135 & .9077 & .9119 & .9735 & .9658 & .9511 & .9354 \\
\multirow{2}{*}{LST-skip} & RSE & (0) & .1843 & .2559 & .3254 & .4643 & .4777 & .4893 & .4950 & .4973 & .0864 & .0931 & .1007 & .1007 & .0226 & .0280 & .0356 & .0449 \\
& CORR & (0) & .9843 & .9690 & .9467 & .8870 & .8721 & .8690 & .8614 & .8588 & .9283 & .9135 & .9077 & .9119 & .9735 & .9658 & .9511 & .9354 \\
\multirow{2}{*}{TPA-LSTM} & RSE & (3) & .1803 & .2347 & .3234 & .4389 & .4487 & .4658 & .4641 & .4765 & .0823 & .0916 & .0964 & .1006 & \textbf{.0174} & \textbf{.0241} & .0341 & \secondbest{.0444} \\
& CORR & (4)& .9850 & .9742 & .9487 & .9081 & .8812 & .8717 & .8717 & .8629 & .9439 & .9337 & .9250 & .9133 & \textbf{.9790} & \textbf{.9709} & \textbf{.9564} & \textbf{.9381} \\
\multirow{2}{*}{MTGNN} & RSE & (5) & .1778 & .2348 & .3109 & .4270 & .4162 & .4754 & \textbf{.4461} & \textbf{.4535} & .0745 & \textbf{.0878} & \secondbest{.0916} & \textbf{.0953} & .0194 & .0259 & .0349 & .0456 \\
& CORR & (6) & .9852 & .9726 & .9509 & .9031 & .8963 & .8667 & \textbf{.8794} & \textbf{.8810} & .9474 & .9316 & .9278 & .9234 & \secondbest{.9786} & \secondbest{.9708} & \secondbest{.9551} & \secondbest{.9372} \\
\midrule
\midrule
\multirow{2}{*}{NE-GNN} & RSE & (2) & .1898 & .2580 & .3472 & .4441 & .4212 & .4586 & .4679 & .4743 & .0762 & .0917 & .0966 & .0994 & .0175 & \secondbest{.0244} & \textbf{.0338} & .0447 \\
& CORR & (0) & .9829 & .9663 & .9367 & .8905 & .8951 & .8748 & .8700 & .8670 & .9494 & .9362 & .9308 & .9262 & .9769 & .9686 & .9535 & .9352 \\
\multirow{2}{*}{FC-GNN} & RSE & \textbf{(12)} & \textbf{.1651} & \textbf{.2202} & \textbf{.2981} & \textbf{.3997} & \textbf{.4057} & \textbf{.4395} & \secondbest{.4624} & \secondbest{.4620} &  \textbf{.0732} & .0907 & \textbf{.0915} & \secondbest{.0979} & \textbf{.0174} & .0245 & .0344 & .0450  \\
& CORR & \textbf{(12)} & \textbf{.9876} & \textbf{.9765} & \textbf{.9551} & \textbf{.9148} & \textbf{.9024} & \textbf{.8850} & \secondbest{.8764} & \secondbest{.8751} & \textbf{.9521} & \textbf{.9404} & .\textbf{9351} & \textbf{.9294} & .9772 & .9685 & .9538 & .9349 \\
\multirow{2}{*}{$\underset{\text{(K=4)}}{\text{BP-GNN}}$} & RSE & (11) & \secondbest{.1704} & \secondbest{.2257} & \secondbest{.3072} & \secondbest{.4050} & \secondbest{.4095} & \secondbest{.4470} & .4640 & .4641 & \secondbest{.0740} & \secondbest{.0898} &  .0940 & .0980 & .0175 & \secondbest{.0244} & \secondbest{.0339} & \textbf{.0442}  \\
& CORR & (10) & \secondbest{.9865} & \secondbest{.9751} & \secondbest{.9522} & \secondbest{.9138}  & \secondbest{.8999} & \secondbest{.8820} & .8744 & .8723 & \secondbest{.9519} & \secondbest{.9396} & \secondbest{.9345} & \secondbest{.9288} & .9769 & .9684 & .9530 & .9360 \\
\bottomrule
\end{tabular}
\vspace{-5pt}
\caption{Benchmark on Solar-Energy, Traffic, Electricity and Exchange-Rate. Root Relative Squared Error (RSE) and Empirical Correlation Coefficient (CORR) are reported for different horizons \{3, 6, 12, 24\}. All results have been averaged over 5 runs. \#Top2 column counts how many metrics in each row are in the top 2 (i.e. bold).  }
\label{tab:electricity_traffic}
\end{center}
\vspace{-5pt}
\end{table*}

\subsection{Single step forecasting} \label{sec:single_step}

In this section, we evaluate our method on the publicly avilable Solar-Energy, Traffic, Electricity and Exchange-Rate datasets. In contrast to METR-LA and PEMS-BAY, these datasets do not contain spatial information from which a graph can be pre-defined, therefore, methods that rely on a known graph are not directly applicable. To ensure comparability, we use the same training settings as \citep{lai2018modeling, shih2019temporal, wu2020connecting} in which the network is trained to predict one time step into the future (single step forecasting) with a given horizon (3, 6, 12, or 24) by minimizing the Mean Absolute Error (MAE). All datasets have been split in 60\%/20\%/20\% for training/val/test respectively. As proposed in \citep{lai2018modeling} we use the Root Relative Squared Error (RSE) and Empirical Correlation Coefficient (CORR) as evaluation metrics (both defined in Appendix \ref{ap:evaluation_metrics}). All our models (FC-GNN, BP-GNN and NE-GNN) contain 2 graph convolutional layers and 128 features in the hidden layers. The timing results have been obtained with batch size 16. Further implementation details are provided in Appendix~\ref{ap:single_step_impl_details}. Results with standard deviations are provided in Appendix~\ref{ap:single_step_std}.


\textbf{Results} in Table \ref{tab:electricity_traffic} are consistent with the previous experiment. FC-GNN outperforms all previous works in most metrics. There is a significant improvement over NE-GNN, which demonstrates that sharing information among nodes is beneficial (except in Exchange-Rate dataset which only contains 8 nodes). BP-GNN performs better than previous methods in most metrics and close to FC-GNN. Regarding the timing results (Table \ref{tab:timing_results2}), BP-GNN is the most efficient graph inference method by a large margin in most datasets. 
The larger the number of nodes in the dataset, the larger the computational improvement of BP-GNN w.r.t. other methods due to its linear scalability when $K=4    $. 
\begin{table}[h!]
\footnotesize
\renewcommand{\arraystretch}{1.0}

\begin{center}
\begin{tabular}{l | cccc  }
\toprule
 & Exchange & Solar & Electricity & Traffic \\
 \midrule
 \# Nodes & 8 & 137 & 321 & 862 \\
 \midrule
 MTGNN & .0062 & .0146 & .0771 & 1.1808\\
 \midrule
 NE-GNN & .0034 & .0059 & .0067 & .0117 \\
 FC-GNN & .0053 & .0109 & .0536 &  .4184\\
 BP-GNN & .0076 & .0076 & .0084 & .0121\\
    \bottomrule
\end{tabular}
\caption{Forward average time in seconds for different methods in Solar-Energy, Traffic, Electricity and Exchange-Rate datasets.}
\vspace{-5pt}
\label{tab:timing_results2}
\vspace{-15pt}
\end{center}
\end{table}
For example, in Solar-Energy (137 nodes), BP-GNN is 1.43 times faster than FC-GNN and 1.92 times faster than MTGNN. Accordingly, in a larger dataset as Electricity (321 nodes) BP-GNN is 6.38 times faster than FC-GNN and 9.18 times faster than MTGNN. In the largest dataset, Traffic (862 nodes), BP-GNN becomes 34.58 times faster than FC-GNN and 97.59 times faster than MTGNN, although in traffic, MTGNN and FC-GNN did not fit in the GPU for a batch of 16 (also because to the $O(N^2)$ complexity), and we had to pass the samples in batches of 2 which eliminates part of the GPU parallelization. Running BP-GNN in batches of 2, would result in it being 7.71 and 18.92 times faster than FC-GNN and MTGNN respectively. Finally, in such small graphs as Exchange Rate, there is no computational benefit in using the bipartite assumption since the number of edges for both the bipartite and the fully connected graphs becomes the same (for K=4 and N=8). 


\subsection{Inferred graph analysis} \label{sec:inferred_graph}
\begin{figure}[t]
\center
\includegraphics[width=1\linewidth]{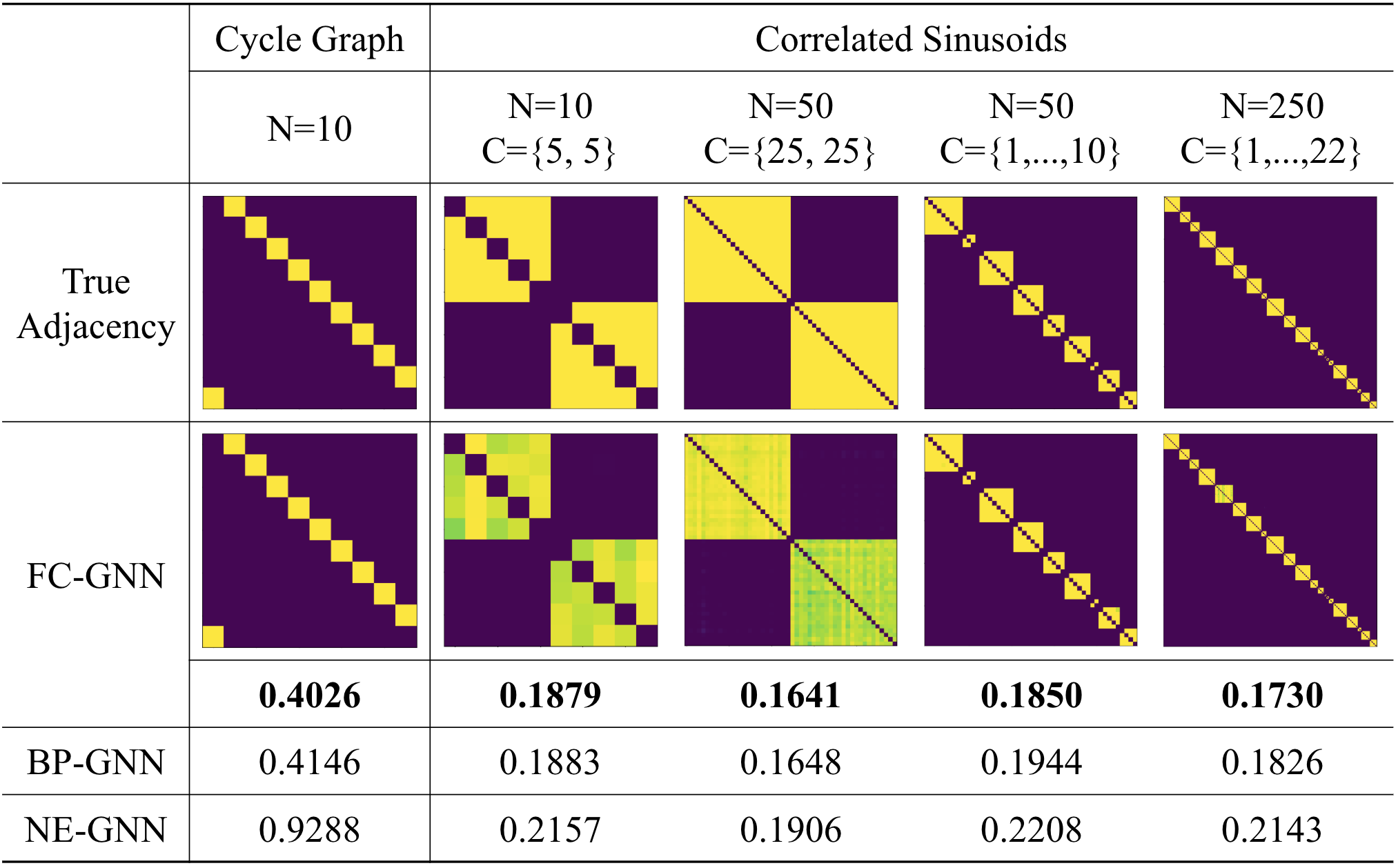}
\vspace{-20pt}
\caption{Inferred adjacency matrices and MAE losses in the proposed synthetic datasets for FC-GNN, BP-GNN and NE-GNN.}
\label{fig:inference_results}
\vspace{-15pt}
\end{figure}

In this section we study the adjacency matrices inferred by FC-GNN. For this purpose, we generated synthetic datasets with different numbers of nodes N and $T=10.000$ timesteps each. (a) "Cycle Graph" dataset samples each series value $\rmx_{i,t}$ from the past $(t-5)$ of another series $(i-1 \mod N)$ from the panel. The resulting adjacency matrix is a directed cycle graph. More formally, the dataset is generated from the following gaussian distribution $\rmx_{i, t} \sim \mathcal{N}(\beta \rmx_{i-1\mod N, t-5}; \sigma^2)$, where $\beta=0.9$ and $\sigma=0.5 ^2$. (b) We name the second dataset "Correlated Sinusoids", inspired by the Discrete Sine Transformation we generate arbitrary signals as the sum of different sinusoids plus gaussian noise. The gaussian noise is independently added to each time series $i$ but the same sinusoid can be shared among different nodes creating strong dependencies. We define different clusters of nodes based on their dependencies, as an example, the first Correlated Sinusoids column in Figure \ref{fig:inference_results} has $N=10$ nodes and two clusters of $5$ nodes each $C=\{5,5\}$, the dataset in the last column has $N=250$ nodes with clusters ranging from size 1 to 22 such that $C=\{1, \dots, 22\}$. Further details and visualizations of these two datasets are provided in Appendix \ref{ap:inferred_graph_datasets}. Both BP-GNN and FC-GNN consist of a single graph convolution from which adjacencies are obtained. We forecast the next time step into the future and optimize the MAE during training. Since our graph inference mechanism is dynamic we average them over 10 timesteps $t$. Further implementation details are described in Appendix \ref{ap:inference_implementation_details}.


\textbf{Results} are reported in Figure \ref{fig:inference_results} for different synthetic datasets which nodes range from $N=10$ to $N=250$. FC-GNN perfectly infers the ground truth adjacencies in all reported datasets and it also achieves the lowest MAE test loss. For the inferred adjacency matrices in BP-GNN, it is harder to develop a visual intuition given the auxiliary nodes, so we omit them from our exposition. However, note that overall accuracy is on par with FC-GNN (specially for datasets with a dense adjacency) and much improved when compared to NE-GNN. So, BP-GNN allows to leverage cross-time series information effectively. Additionally, we also provide visualizations of the inferred graphs in METR-LA in Appendix \ref{ap:inferred_metrla} where we demonstrate the dynamic inference behavior of the graph by plotting the inferred adjacency at different time steps $t$.

\subsection{Choosing the number of auxiliary nodes K} \label{sec:scalability}
\begin{figure}[h!]
\center
\includegraphics[width=0.95\linewidth]{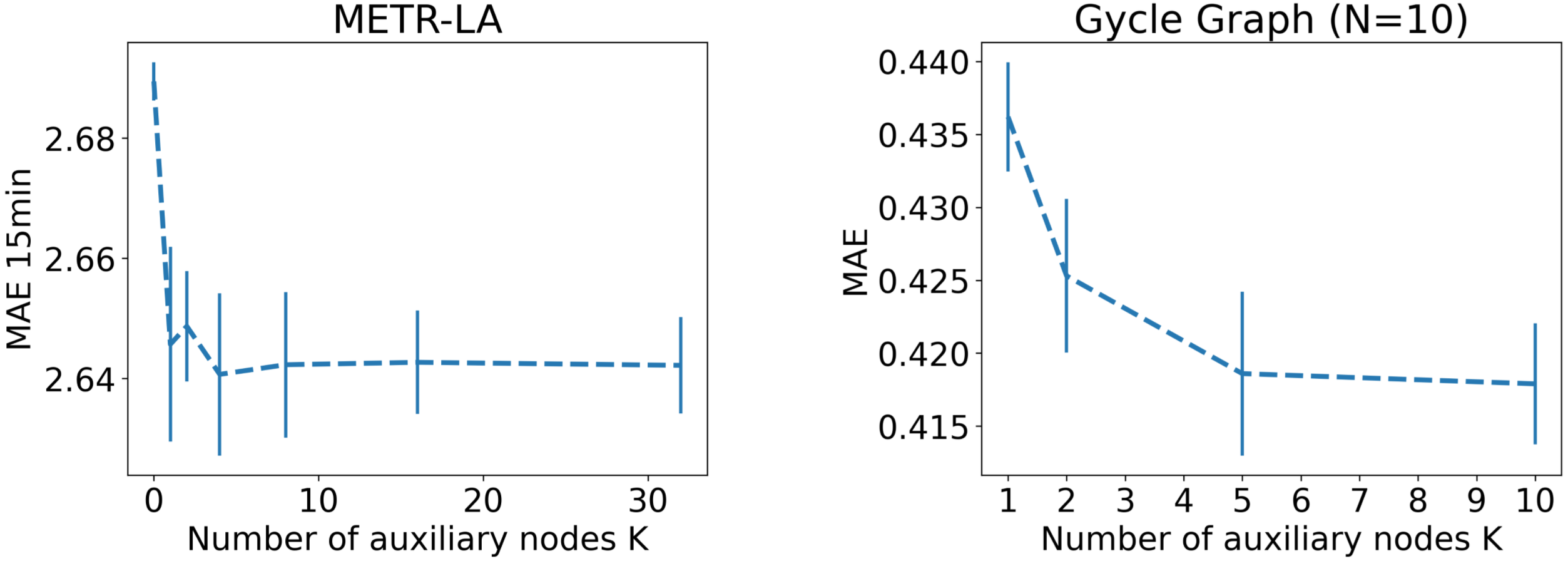}
\vspace{-5pt}
\caption{Left: MAE in METR-LA when sweeping $K$ from 0 to 32 auxiliary nodes (10 runs average). Right: MAE in Cycle Graph N=10, when sweeping $K$ from 1 to 10 (4 runs average).}
\label{fig:sweep_k}
\vspace{-3pt}
\end{figure}

In all previous sections we set a relatively small number of auxiliary nodes ($K=4$) for BP-GNN. In this section we analyze the BP-GNN performance as we sweep over different $K$ values  (Figure \ref{fig:sweep_k}). We noticed that for sparser graph datasets as "Cycle Graph", increasing $K$ benefits more than in less sparse graph datasets. In practice, in previous experiments, we chose $K$ to be relatively small since in real world datasets as METR-LA, small $K$ values already provide competitive accuracy with a good trade-off in complexity.

\vspace{-5pt}

\subsection{Complexity analysis}

\begin{figure}[h!]
\center
\includegraphics[width=0.95\linewidth]{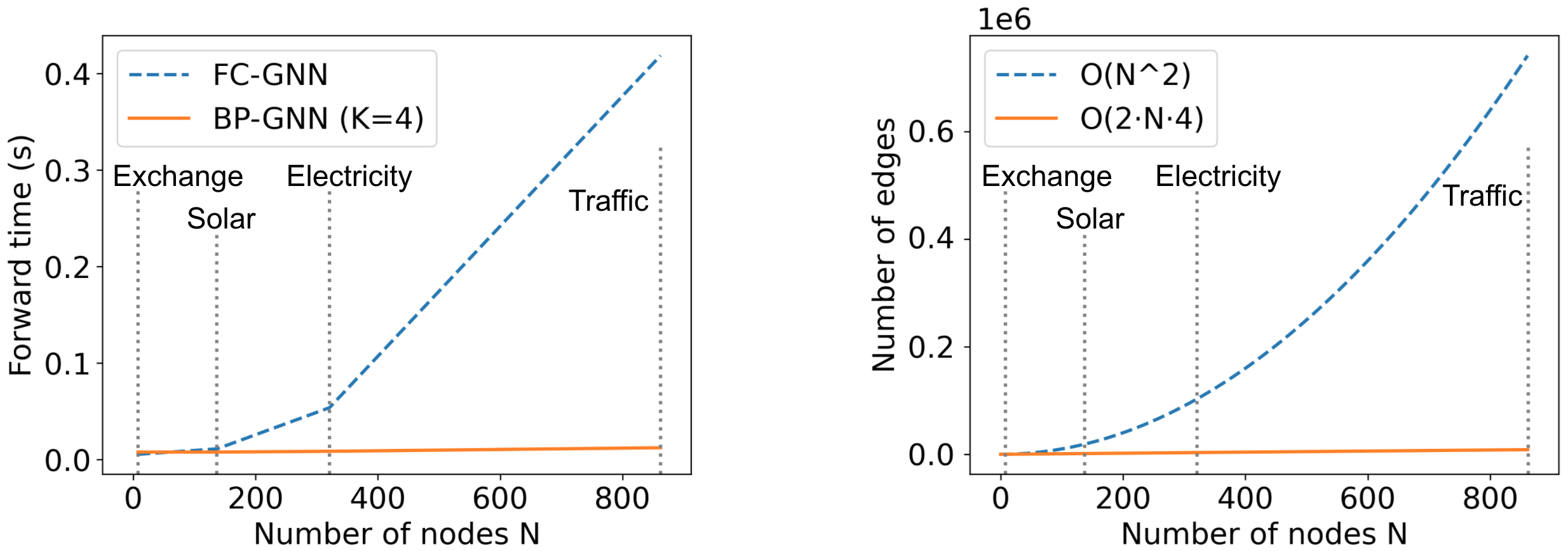}
\vspace{-5pt}
\caption{Left: Forward time in seconds for FC-GNN and BP-GNN on different datasets. Right: Scalability of the number of edges w.r.t. the number of nodes $N$.}
\label{fig:complexity}
\vspace{-10pt}
\end{figure}

In Figure \ref{fig:complexity} we provide a visual comparison of the scalability of FC-GNN and BP-GNN (K=4). The right plot shows the vast difference in the number of edges between both methods as we increase the number of nodes. The left plot shows how this translates to the forward time in seconds in the experiments from Section \ref{sec:single_step}.
\vspace{-10pt}

%% file: sections/conclusions.tex

\section{Conclusions}
We presented a novel approach for multi-variate time series forecasting which lends itself to easy integration into existing univariate approaches by ways of a graph neural network (GNN) component. This GNN infers a latent graph in the space of the embeddings of the univariate time series and can be inserted in many neural-network based forecasting models. We show that this additional graph/dependency structure improves forecasting accuracy both in general and in particular when compared to the state of the art. We alleviate typical scalability concerns for GNNs via allowing the introduction of auxiliary nodes in the latent graph which we construct as a bipartite graph that bounds the computational complexity. We show that this leads to computationally favorable properties as expected while the trade-off with forecasting accuracy is manageable. 

%% file: sections/appendix.tex
\newpage
\appendix 

\section{Model details}
\label{ap:implementation_details} 
 
\subsection{BP-GNN Matrices} \label{ap:bg_gnn}
Here, we define the adjacency matrices corresponding to the two message passing steps of the bipartite graph for $N$ time series nodes and $K$ auxiliary nodes (assuming all inference parameters $\alpha_{ij} = 1$). We also present the product of both matrices $\tilde{A} = A_2A_1$. The first $N$ rows of $\tilde{A}$ can be interpreted as the sum of all paths that communicate the $N$ time series nodes among them after applying the two message passing updates $A_2$ and $A_1$. 
\begin{equation*}
\small
  A_1 =  \begin{vmatrix} 
    0_{N \times N} & 0_{N \times K}   \\
   1_{K \times N} & 0_{K \times K} \\
   \end{vmatrix} 
,\qquad
  A_2 =  \begin{vmatrix} 
  0_{N \times N} & 1_{N \times K}   \\
   0_{K \times N} & 0_{K \times K} \\
   \end{vmatrix} 
   ,\qquad
      \tilde{A} = A_2A_1 =  \begin{vmatrix} 
   K_{N \times N} & 0_{N \times K}   \\
   0_{K \times N} & 0_{K \times K} \\
   \end{vmatrix} 
\end{equation*}

\subsection{Architecture choices $\phi_e$, $\phi_\alpha$, $\phi_h$, $f_{enc}$ and $f_{dec}$}
\label{ap:architecture_choices}

In this section we will use the shortcut $\text{MLP}_{res}$ for a one layer MLP with a residual connection which we define as:

\quad Input $\xrightarrow{}$ \{LinearLayer(nf, nf) $\xrightarrow{}$ Swish() $\xrightarrow{}$ LinearLayer(nf, nf) $\xrightarrow{}$ Addition(Input) \} $\xrightarrow{}$ Output
 
Where "nf" represents the number of features.
 
\textbf{Decoder} $f_{dec}$

Given the above $\text{MLP}_{res}$ definition, the decoder consists of one residual MLP  followed by a linear layer:

\quad $\hat{\rmz}_i$ $\xrightarrow{}$ \{ $\text{MLP}_{res}$ $\xrightarrow{}$ LinearLayer(nf, out\_dim)\} $\xrightarrow{}$ $\hat{\rmx}_{i,t+1:T}$

\textbf{Encoder} $f_{enc}$

The encoder for METR-LA and PEMS-BAY consists of a linear layer and two consecutive $\text{MLP}_{res}$. Notice the overall structure can be considered an MLP with residual connections among some of its layers.

\quad $\rmx_{i, t_0:t}$ $\xrightarrow{}$ \{ LinearLayer(in\_dim, nf) $\xrightarrow{}$ $\text{MLP}_{res}$ $\xrightarrow{}$ $\text{MLP}_{res}$ $\xrightarrow{}$ concatenate($\rmc_i$)\} $\xrightarrow{}$ $\rmz_i$

In the other datasets (Solar-Energy, Traffic, Electricity and Exchange-Rate) we used the following Convolutional Neural Network as an encoder:

\quad $\rmx_{i, t_0:t}$ $\xrightarrow{}$ \{conv1d(in\_dim, nf, kernel\_size, stride) $\xrightarrow{}$ $\text{CNN}_{res}(\text{nf})$ 

\qquad \quad $\xrightarrow{}$ conv1d(nf, 2nf, kernel\_size, stride) $\xrightarrow{}$ $\text{CNN}_{res}(2\text{nf})$

\qquad \quad $\xrightarrow{}$ conv1d(2nf, 4nf, kernel\_size, stride) $\xrightarrow{}$ $\text{CNN}_{res}(4\text{nf})$

\qquad \quad $\xrightarrow{}$ conv1d(4nf, out\_dim, 1, 1) $\xrightarrow{}$ concatenate($\rmc_i$) \} $\xrightarrow{}$ $\rmz_i$

Where $\text{CNN}_{res}$ follows the same architecture than $\text{MLP}_{res}$ but replacing the Linear Layers by Conv1d layers with stride=1 and kernel\_size=1. In other words, both architectures are equivalent since a CNN with kernel size and stride 1 is equivalent to an MLP that broadcasts over the batch size and sequence length. Formally, we write $\text{CNN}_{res}$ as:

{\small\quad Input $\xrightarrow{}$ \{conv1d(nf, nf/2, 1, 1) $\xrightarrow{}$ Swish() $\xrightarrow{}$ conv1d(nf/2, nf, 1, 1) $\xrightarrow{}$ Addition(Input) \} $\xrightarrow{}$ Output}

\textbf{Edge update} $\phi_e$

Consists of two layers MLP. We divide the number of features by two in the intermediate layer for efficiency

\quad $[\rmh_i, \rmh_j]$ $\xrightarrow{}$ \{LinearLayer(2$\cdot$nf, nf/2) $\xrightarrow{}$ Swish() $\xrightarrow{}$ LinearLayer(nf/2, nf) $\xrightarrow{}$ Swish() \} $\xrightarrow{}$ $\rmm_{ij}$. 

\textbf{Node update} $\phi_h$

\quad [$\rmh_i^{l}$, $\rmm_i$] $\xrightarrow{}$ \{LinearLayer(2$\cdot$ nf, nf) $\xrightarrow{}$ Swish() $\xrightarrow{}$ LinearLayer(nf, nf) $\xrightarrow{}$ Addition($\rmh^l_i$) \} $\xrightarrow{}$ $\mathbf{h}^{l+1}_i$ 

\textbf{Edge inference} $\phi_\alpha$

\quad [$\rmm_{ij}$] $\xrightarrow{}$ \{LinearLayer(nf, 1) $\xrightarrow{}$ Sigmoid() \} $\xrightarrow{}$ $\alpha_{ij}$

\subsection{Linear Graph Convolutional Layer} \label{ap:linear_graph}
We can define a linear graph convolutional layer $H^{l+1} = \mathrm{GCL}(H, A)$ as:
\begin{align}
    Z &= AH^l\theta_e + IH^l\theta_h \\
   H^{l+1} &= \sigma(Z)
\end{align}

Where $A \in \{0, 1 \}^{N \times N}$ is the adjacency matrix of the graph. $H \in \mathbb{R}^{N \times \text{nf}}$ a matrix of node features with $N$ nodes and "nf" features per node. $I \in \{0, 1 \}^{N \times N}$ is an identity matrix and $\theta_e, \theta_h \in \mathbb{R}^{\text{nf} \times \text{nf}}$ the learnable parameters.

\section{METR-LA and PEMS-BAY} \label{ap:metr_pems}
METR-LA is a traffic dataset collected from the highway of Los Angeles. It contains 207 nodes, a sampling resolution of 5 minutes and 34,272 samples per node (i.e. length of each time series).

PEMS-BAY is also a traffic dataset with 325 nodes located in Bay Area. The sampling resolution is also 5 minutes and it contains 52,116 samples per time series / node.

Both datasets METR-LA and PEMS-BAY can be download from the following link:
\url{https://github.com/liyaguang/DCRNN}

\begin{table}[h]
\footnotesize
\begin{center} 
\begin{tabular}{l | cc ccc  }
\toprule
 & (N) \#Nodes & \# Samples & Resolution & Context length & Pred. length\\
 \midrule
 METR-LA & 207 & 34,272 & 5 min & 12 & 12 \\
PEMS-BAY & 325 & 52,116 & 5 min & 12 & 12 \\
    \bottomrule
\end{tabular}
\vspace{-5pt}
\caption{METR-LA and PEMS-BAY specifications.}
\end{center}
\vspace{-15pt}
\end{table}

\subsection{Implementation details} \label{ap:impl_metr_pems}
In this experiment (METR-LA and PEMS-BAY datasets) we used the exact same training configuration as \citep{shang2021discrete}. Given the whole training time series panel of dimensionality (N $\times$ Length), where Length is the number of samples of the training partition, we construct the input to the network by uniformly slicing windows $\rmx_{t_0:t} \in \mathbb{R}^{N\times 12}$ of length 12. The ground truth labels are obtained in the same way, by slicing the next 12 time steps $\rmx_{t:t+12} \in \mathbb{R}^{N\times 12}$. The same slicing process is done in validaton and test with their respective partitions. The sliced windows are uniformly distributed through the whole time series panel. 

All our models (FC-GNN, BP-GNN, NE-GNN) have been trained with Adam optimizer, batch size 16 and 2 layers in the GNN module. The number of features "nf" in the hidden layers is 64.  The learning rates and number of epochs are provided in the following Table \ref{tab:hyperparams_1}.

\begin{table}[h]
\footnotesize

\begin{center} 
\begin{tabular}{l | c | cccc  } 
\toprule
 & & Learning Rate & Decay at epochs & Decay factor & Epochs \\
\midrule
\multirow{3}{*}{METR-LA}&  NE-GNN & $2 \cdot 10^{-3}$ & [20, 30, 40] & 10 & 200 \\
& FC-GNN & $5 \cdot 10^{-3}$ & [20, 30, 40] & 10 & 200 \\
& BP-GNN & $2 \cdot 10^{-3}$ & [20, 30, 40] & 10 & 200 \\
\midrule
\multirow{3}{*}{PEMS-BAY}&  NE-GNN & $2 \cdot 10^{-4}$ & [50] & 10 & 300\\
& FC-GNN & $2 \cdot 10^{-4}$ & [50] & 10 & 300 \\
& BP-GNN & $2 \cdot 10^{-4}$ & [50] & 10 & 300 \\
\bottomrule
\end{tabular}
\caption{Learning rates for METR-LA and PEMS-BAY datasets }
\label{tab:hyperparams_1}
\end{center}
\end{table}

Time results have been run for batch size 16 in a Tesla V100-SXM GPU. All hyperparameters have been tuned in the validation partition using the following search spaces. The learning rate search space was lr $\in$ $\{1\cdot 10^{-4}, 2\cdot 10^{-4}, 5\cdot 10^{-4}, 1\cdot 10^{-3}, 2\cdot 10^{-3}, 5\cdot 10^{-3} \}$. The number of features per layer "nf" was chosen among nf $\in \{32, 64, 128\}$. The number of layers was chosen among $\{1, 2, 3, 4\}$. The learning rate decay and the decay factor was not modified and left the same as in the original code from \citep{shang2021discrete}. The number of epochs was chosen large enough to do early stopping in the validation partition when the validation loss stops decreasing.

\subsection{Baselines}

Regarding the baselines reported in this experiment, LDS \citep{franceschi2019learning} and STGCN \citep{yu2017spatio} are not evaluated on METR-LA and PEMS-BAY in their original papers, therefore, for STGCN we used the results provided in the WaveNet paper \citep{wu2019graph} which were computed by the same WaveNet authors and for LDS we used the results provided in the GTS work \citep{shang2021discrete}.
For the DCRNN timing results we used the unofficial Pytorch implementation \url{https://github.com/chnsh/DCRNN_PyTorch}.

\subsection{METR-LA and PEMS-BAY results with standard devitation} \label{ap:metr_pems_std}
In this subsection we report the standard deviations for METR-LA and PEMS-BAY results.

\begin{table}[h]
\scriptsize
\setlength{\tabcolsep}{3pt}
\begin{center}
\begin{tabular}{l | ccc | ccc | ccc | }
\toprule
 & \multicolumn{9}{c|}{METR-LA} \\
 & \multicolumn{3}{c}{15 min} & \multicolumn{3}{c}{30 min} & \multicolumn{3}{c|}{60 min}\\
 \midrule
  & \tiny MAE & \tiny RMSE & \tiny MAPE & \tiny MAE & \tiny RMSE & \tiny MAPE & \tiny MAE & \tiny RMSE & \tiny MAPE  \\

     \midrule
       \textit{Ablation study} &    \\
 NE-GNN \textit{(w/o id)} & 2.80 $\pm$ .01 & 5.73 $\pm$ .02 & 7.50\% $\pm$ .03 & 3.40 $\pm$ .00 & 7.15 $\pm$ .02  & 9.74\% $\pm$ .12 & 4.22 $\pm$ .01 & 8.79 $\pm$ .04 & 13.06\% $\pm$ .26 \\
 FC-GNN \textit{(w/o id)} & 2.77 $\pm$ .00 & 5.65 $\pm$ .01 & 7.39\% $\pm$ .06 & 3.36 $\pm$ .00 & 7.02 $\pm$ .02 & 9.59\% $\pm$ .06 & 4.14 $\pm$ .01 & 8.64 $\pm$ .06 & 12.70\% $\pm$ .11\\
 NE-GNN & 2.69 $\pm$ .01 & 5.57 $\pm$ .04 & 7.21\% $\pm$ .01 & 3.14 $\pm$ .01 & 6.74 $\pm$ .05 & 9.01\% $\pm$ .09 & 3.62 $\pm$ .01 & 7.88 $\pm$ .05 & 10.94\% $\pm$ .07 \\
 \midrule
 \textit{Our models} & \\
 FC-GNN & 2.60 $\pm$ .02 & 5.19 $\pm$ .06 & 6.78\% $\pm$ .12 & 2.95 $\pm$ .02 & 6.15 $\pm$ .08 & 8.14\% $\pm$ .16 & 3.35 $\pm$ .03 & 7.14 $\pm$ .09 & 9.73\% $\pm$ .27 \\
  BP-GNN (K=4) & 2.64 $\pm$ .01 & 5.37 $\pm$ .03 & 7.07\% $\pm$ .08 & 3.02 $\pm$ .02 & 6.42 $\pm$ .05 & 8.46\% $\pm$ .08 &  3.40 $\pm$ .02 & 7.32 $\pm$ .05 & 9.91\% $\pm$ .19 \\
    \bottomrule
\end{tabular}
\caption{METR-LA results including standard deviations.}
\vspace{-10pt}
\end{center}
\end{table}

\begin{table}[h]
\scriptsize
\setlength{\tabcolsep}{3pt}
\begin{center}
\begin{tabular}{l | ccc | ccc | ccc | }
\toprule
 & \multicolumn{9}{c|}{PEMS-BAY} \\
 & \multicolumn{3}{c}{15 min} & \multicolumn{3}{c}{30 min} & \multicolumn{3}{c|}{60 min} \\
 \midrule
  & \tiny MAE & \tiny RMSE & \tiny MAPE & \tiny MAE & \tiny RMSE & \tiny MAPE & \tiny MAE & \tiny RMSE & \tiny MAPE\\

     \midrule
       \textit{Ablation study} &    \\
 NE-GNN \textit{(w/o id)} & 1.40 $\pm$ .00 & 3.03 $\pm$ .01 & 2.92\% $\pm$ .02 & 1.85 $\pm$ .00 & 4.25 $\pm$ .02 & 4.21\% $\pm$ .04 & 2.39 $\pm$ .01  & 5.50 $\pm$ .03 & 5.93\% $\pm$ .07  \\
 FC-GNN \textit{(w/o id)} & 1.39 $\pm$ .00 & 3.00 $\pm$ .01 & 2.88\% $\pm$ .01 & 1.82 $\pm$ .00 & 4.18 $\pm$ .01 & 4.11\% $\pm$ .01 & 2.32 $\pm$ .01 & 5.35 $\pm$ .03 & 5.71\% $\pm$ .02 \\
 NE-GNN & 1.36 $\pm$ .00 & 2.88 $\pm$ .01 & 2.86\% $\pm$ .02 & 1.72 $\pm$ .01 & 3.91 $\pm$ .03 & 3.93 \% $\pm$ .05 & 2.07 $\pm$ .01 & 4.79 $\pm$ .04 & 5.04\% $\pm$ .06 \\
 \midrule
 \textit{Our models} & \\
 FC-GNN & 1.33 $\pm$ .00 & 2.82 $\pm$ .01 & 2.79\% $\pm$ .03 & 1.65 $\pm$ .00 & 3.75 $\pm$ .01 & 3.72\% $\pm$ .06 & 1.93 $\pm$ .01 & 4.46 $\pm$ .02 & 4.53\% $\pm$ .06    \\
  BP-GNN (K=4) & 1.33 $\pm$ .00 &  2.82 $\pm$ .01 & 2.80\% $\pm$ .02 & 1.66 $\pm$ .01 & 3.78 $\pm$ .01 & 3.75\% $\pm$ .03 & 1.94 $\pm$ .01 & 4.46 $\pm$ .02 & 4.57\% $\pm$ .04 \\
    \bottomrule
\end{tabular}
\caption{PEMS-BAY results including standard deviations.}
\vspace{-10pt}
\end{center}
\end{table}

\section{Single step forecasting} \label{ap:single_step_forecasting}

Electricity, Solar-Energy, Electricity and Exchange Rate datasets can be downloaded from the following link: \url{https://github.com/laiguokun/multivariate-time-series-data}

\begin{table}[h]
\footnotesize
\setlength{\tabcolsep}{4pt}
\begin{center} 
\begin{tabular}{l | cc ccc  }
\toprule
 & \#Nodes & \# Samples & Resolution & Context length & Pred. length\\
 \midrule
Solar-Energy & 137 & 52,560 & 10 min & 168 & 1\\
Traffic & 862 & 17,544 & 1 hour & 168 & 1\\
Electricity & 321 & 26,304 & 1 hour & 168 & 1\\
Exchange-Rate & 8  &  7,588 & 1 day & 168 & 1\\
    \bottomrule
\end{tabular}
\vspace{-5pt}
\caption{Dataset specifications.}
\end{center}
\vspace{-15pt}
\end{table}

\subsection{Implementation details} \label{ap:single_step_impl_details}
In this experiment we used the exact same training process as \citep{wu2020connecting}. Similarly to METR-LA and PEMS-BAY, the sample construction process, consists of windows uniformly sliced from the time series panel and inputted to the network. Specifically, in this experiment, given the training time series panel of dimensionality (N $\times$ Length), where Length is the number of samples of the training partition, we slice windows $\rmx_{t_0:t} \in \mathbb{R}^{N\times 168}$ of length 168 which are the input to our network. In this experiment, we only forecast one time step into the future such that the length of the forecasts and ground truth labels is 1.

All our models (FC-GNN, BP-GNN, NE-GNN) have been trained with Adam optimizer. The number of features "nf" in the hidden layers is 128, the number of layers in the GNN is 2. The encoder network in this experiment is a Convolutional Neural Network previously described in Appendix \ref{ap:implementation_details}. Following, we provide a table with the the different learning rates, number of epochs and batch sizes for all datasets. All models NE-GNN, FC-GNN and BP-GNN were trained with the same training parameters.

\begin{table}[h]
\footnotesize

\begin{center} 
\begin{tabular}{l | ccc  } 
\toprule
 & Learning Rate & Batch Size & Epochs \\
\midrule
Solar-Energy & $5\cdot 10 ^{-4}$ & $4$ & $30$ \\
Traffic & $2\cdot 10 ^{-4}$ & $2$ & $50$ \\
Electricity & $2\cdot 10 ^{-4}$ & $4$ & $80$ \\
Exchange-Rate & $1\cdot 10 ^{-4}$ & $4$ & $100$ \\

\bottomrule
\end{tabular}
\caption{Table of hyperparameters for Solar-Energy, Traffic, Electricity and Exchange-Rate. }
\label{tab:hyperparams_3}
\end{center}
\end{table}

All hyperparameters have been tuned in the validation partition using the following search spaces. The learning rate search space was lr $\in$ $\{5\cdot 10^{-5}, 1\cdot 10^{-4}, 2\cdot 10^{-4}, 5\cdot 10^{-4}, 1\cdot 10^{-3}, 2\cdot 10^{-3}\}$. The number of features nf was chosen among \{32, 64, 128\}, the number of layers chosen among \{1, 2, 3, 4, 8\} by choosing between a trade-off of accuracy and efficiency. The number of epochs was chosen large enough such that the validation loss would stop decreasing. The batch size was set to 4, except for Traffic where it was reduced to 2 in order to fit in memory the more expensive explored configurations.

\subsection{Evaluation metrics} \label{ap:evaluation_metrics}
The evaluation metrics for this experiments are exactly the same as for \cite{lai2018modeling,shih2019temporal, wu2020connecting}. The following equations are extracted from \cite{lai2018modeling}.
The Root Relative Squared Error (RSE) is defined as:

$$
R S E=\frac{\sqrt{\sum_{(i, t) \in \Omega_{T e s t}\left(\rmx_{i t}-\hat{\rmx}_{i t}\right)^{2}}}}{\sqrt{\sum_{(i, t) \in \Omega_{T e s t}\left(\rmx_{i t}-\operatorname{mean}(\rmx)\right)^{2}}}}
$$

The Empirical Correlation Coefficient (CORR) is defined as:

$$
\operatorname{CORR}=\frac{1}{N} \sum_{i=1}^{N} \frac{\sum_{t}\left(\rmx_{i t}-\operatorname{mean}\left(\rmx_{i}\right)\right)\left(\hat{\rmx}_{i t}-\operatorname{mean}\left(\hat{\rmx}_{i}\right)\right)}{\sqrt{\sum_{t}\left(\rmx_{i t}-\operatorname{mean}\left(\rmx_{i}\right)\right)^{2}\left(\hat{\rmx}_{i t}-\operatorname{mean}\left(\hat{\rmx}_{i}\right)\right)^{2}}}
$$

Where $\rmx, \hat{\rmx} \in \mathbb{R}^{N \times T}$.

\subsection{Single step results with standard devitation} \label{ap:single_step_std}
In this subsection we report the standard deviations for Solar-Energy, Traffic, Electricity and Exchange-Rate results.

\begin{table}[h]
\scriptsize
\setlength{\tabcolsep}{2 pt}
\begin{center}
\begin{tabular}{ll | cccc | cccc }
\toprule
\multicolumn{2}{c|}{Dataset} & \multicolumn{4}{|c|}{Solar-Energy} & \multicolumn{4}{|c}{Traffic} \\
\midrule
\multicolumn{2}{c|}{} & \multicolumn{4}{c}{Horizon} & \multicolumn{4}{|c}{Horizon} \\
Methods & Metrics &
3 & 6 & 12 & 24 &
3 & 6 & 12 & 24 \\
\midrule
\multirow{2}{*}{NE-GNN} & RSE & .1898 $\pm$ .0018 & .2580 $\pm$ .0013 & .3472 $\pm$ .0059 & .4441 $\pm$ .0083 & .4212 $\pm$ .0007 & .4586 $\pm$ .0017 & .4679 $\pm$ .0031 & .4743 $\pm$ .0036 \\
& CORR & .9829 $\pm$ .0003 & .9663 $\pm$ .0003 & .9367 $\pm$ .0025 & .8905 $\pm$ .0052 & .8951 $\pm$ .0005 & .8748 $\pm$ .0007 & .8700 $\pm$ .0015 & .8670 $\pm$ .0013 \\
\multirow{2}{*}{FC-GNN} & RSE & .1651 $\pm$ .0006 & .2202 $\pm$ .0020 & .2981 $\pm$ .0035 & .3997 $\pm$ .0047 & .4057 $\pm$ .0012 & .4395 $\pm$ .0049 & .4624 $\pm$ .0021 & .4620 $\pm$ .0033 \\
& CORR & .9876 $\pm$ .0002 & .9765 $\pm$.0003 & .9551 $\pm$ .0011 & .9148 $\pm$ .0024 & .9024 $\pm$ .0006 & .8850 $\pm$ .0004 & .8764 $\pm$ .0015 & .8751 $\pm$ .0007 \\
\multirow{2}{*}{$\underset{\text{(K=4)}}{\text{BP-GNN}}$} & RSE & .1704 $\pm$ .0017 & .2257 $\pm$ .0020 & .3072 $\pm$ .0095 & .4050 $\pm$ .0082 & .4095 $\pm$ .0012 & .4470 $\pm$ .0062 & .4640 $\pm$ .0033 & .4641 $\pm$ .0024 \\
& CORR & .9865 $\pm$ .0002 & .9751 $\pm$ .0005 & .9522 $\pm$ .0033 & .9138 $\pm$ .0024  & .8999 $\pm$ .0005 & .8820 $\pm$ .0007 & .8744 $\pm$ .0015 & .8723 $\pm$ .0013  \\
\bottomrule
\end{tabular}
\caption{Solar-Energy and Traffic results including standard deviations.}
\end{center}
\end{table}

\begin{table}[h]
\scriptsize
\setlength{\tabcolsep}{2pt}
\begin{center}
\begin{tabular}{ll | cccc | cccc }
\toprule
\multicolumn{2}{c|}{Dataset} &
\multicolumn{4}{|c|}{Electricity} &
\multicolumn{4}{|c}{Exchange-Rate} \\
\midrule
\multicolumn{2}{c|}{} & \multicolumn{4}{c}{Horizon} & \multicolumn{4}{|c}{Horizon}  \\
Methods & Metrics &
3 & 6 & 12 & 24 &
3 & 6 & 12 & 24 \\
\midrule
\multirow{2}{*}{NE-GNN} & RSE & .0762 $\pm$ .0007 & .0917 $\pm$ .0029 & .0966 $\pm$ .0018 & .0994 $\pm$ .0019 & .0175 $\pm$ .0002 & .0244 $\pm$ .0002 & .0338 $\pm$ .0006 & .0447 $\pm$ .0008 \\
& CORR & .9494 $\pm$ .0006 & .9362 $\pm$ .0009 & .9308 $\pm$ .0006 & .9262 $\pm$ .0006 & .9769 $\pm$ .0001 & .9686 $\pm$ .0005 & .9535 $\pm$ .0003 & .9352 $\pm$ .0003 \\
\multirow{2}{*}{FC-GNN} & RSE &   .0732 $\pm$ .0007 & .0907 $\pm$ .0041 & .0915 $\pm$ .0026 & .0979 $\pm$ .0030 & .0174 $\pm$ .0001 & .0245 $\pm$ .0002 & .0344 $\pm$ .0010 & .0450 $\pm$ .0013  \\
& CORR &  .9521 $\pm$ .0008 & .9404 $\pm$ .0008 & .9351 $\pm$ .0007 & .9294 $\pm$ .0006 & .9772 $\pm$ .0001 & .9685 $\pm$ .0004 & .9538 $\pm$ .0008 & .9349 $\pm$ .0007 \\
\multirow{2}{*}{$\underset{\text{(K=4)}}{\text{BP-GNN}}$} & RSE &  .0740 $\pm$ .0010 & .0898 $\pm$ .0041 &  .0940 $\pm$ .0025 & .0980 $\pm$ .0019 & .0175 $\pm$ .0001 & .0244 $\pm$ .0003 & .0339 $\pm$ .0004 & .0442 $\pm$ .0005  \\
& CORR &  .9519 $\pm$ .0004 & .9396 $\pm$ .0007 & .9345 $\pm$ .0007 & .9288 $\pm$ .0001 & .9769 $\pm$ .0003 & .9684 $\pm$ .0002 & .9530 $\pm$ .0007 & .9360 $\pm$ .0011 \\
\bottomrule
\end{tabular}
\caption{Electricity and Exchange-Rate results including standard deviations.}
\end{center}
\end{table}

\section{Inferred Graph Analysis} \label{ap:inferred_graph_analysis}
\subsection{Datasets} \label{ap:inferred_graph_datasets}

In the Inferred graph analysis experiment we presented two synthetic datasets, "Cycle Graph" and "Correlated Sinusoids". Next, we provide a more detailed explanation on how these datasets have been generated:

\begin{itemize}
    \item Cycle Graph: This dataset consists of a panel of N=10 time series of length T=10.000, where each time series $i$ at time step $t$ has been sampled from the past $t-5$ of another time series $i-1 \mod N$ from the same panel. The resulting multivariate time series adjacency matrix is a Cyclic Directed Graph where each variable in the panel depends on the previously indexed one. Formally the generation process is written as
    \begin{equation}
        \rmx_{i, t} \sim \mathcal{N}(\beta \rmx_{(i-1\mod N), t-5} ; \sigma^2)
    \end{equation}   
    Where $\beta=0.9$ and $\sigma=0.5$. 
    
    \item Correlated Sinusoids: Motivated by the Discrete Sine Transformation. This dataset consists time series of length $T=10.000$, each time series generated as the weighted sum of different sinusoids with different frequencies and amplitudes. More formally we define a time series $\rmx_i$ as:
    \begin{equation}
        \rmx_i = \sum_{k=1}^M B_{i,m}\sin(2\pi w_{i,m} t) + \epsilon_{i,t}
    \end{equation}
    
    Where $B_{i,k}$ is sampled from a Uniform distribution $\tilde{B}_{i,m} \sim \mathcal{U}(0, 1)$ and further normalized $B_{i,m} = \frac{\tilde{B}_{i,m}}{\sum_k \tilde{B}_{i,m}}$.  $w_{i,m}$ is also sampled from a uniform distribution $w_{i,m} \sim \mathcal{U}(0, 0.2)$ and
    $\epsilon_{i,t}$ is sampled from a Gaussian distribution $\epsilon_{i,t} \sim \mathcal{N}(0, 0.2^2)$. Finally, $B_{i,m}$ and $w_{i,m}$ are shared among different variables (i.e. time series) in the pannel defining dependencies among them. We choose $M=3$.

\end{itemize}

\begin{figure}[h!]
\center
\includegraphics[width=0.9\linewidth]{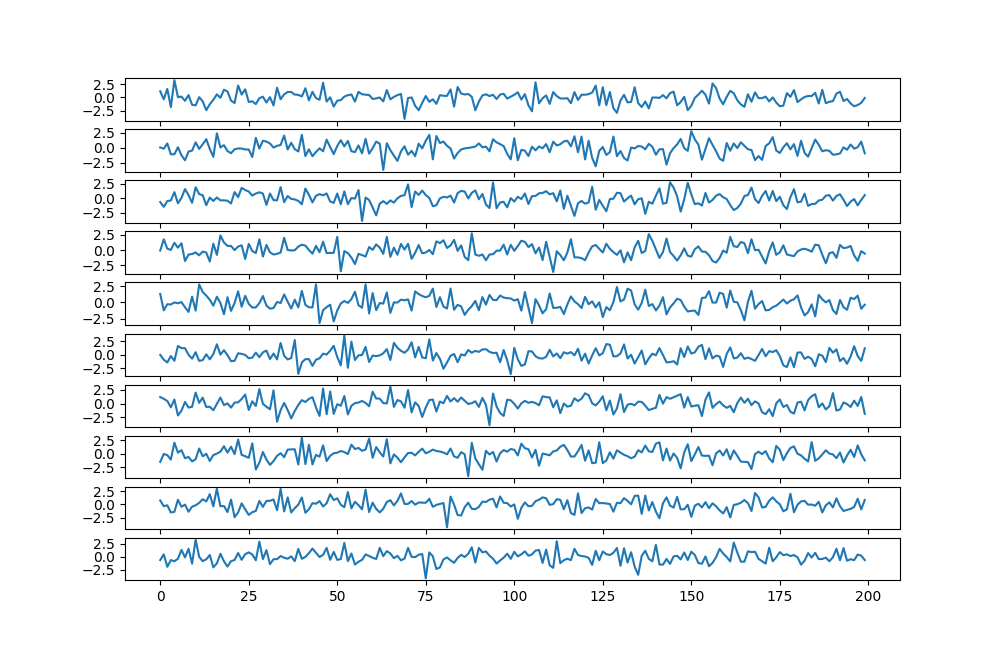}
\caption{First 200 timesteps of the Cycle Graph Gaussian dataset.}
\label{fig:cycle_graph_gaussians}
\end{figure}

\begin{figure}[h!]
\center
\includegraphics[width=0.9\linewidth]{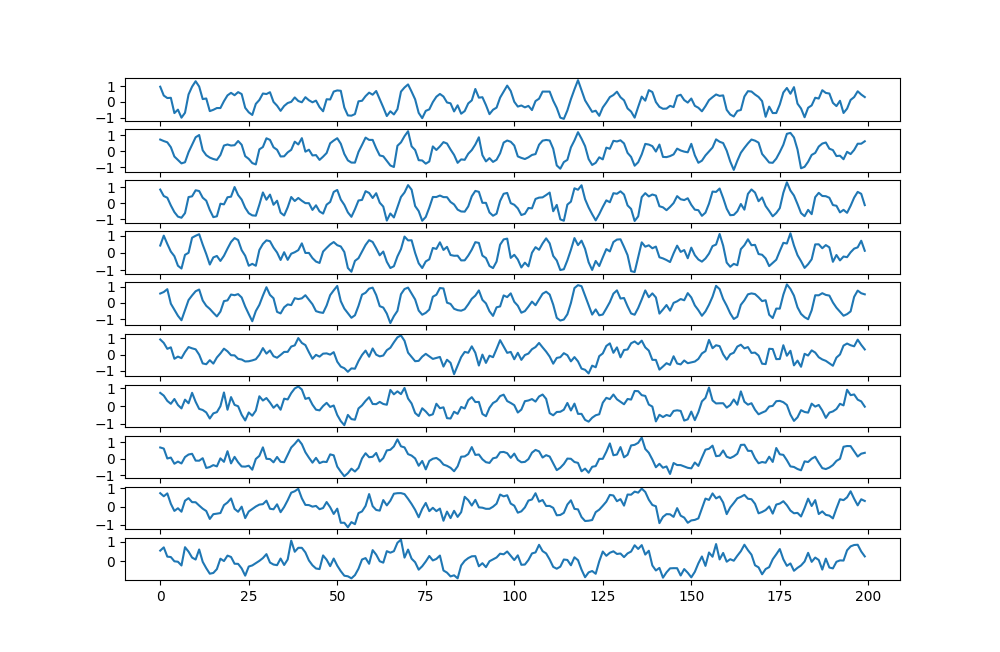}
\caption{First 200 timesteps of Noisy Sinusoids dataset.}
\label{fig:noisy_sinusoids}
\end{figure}

The time series in both datasets have been splitted in train/val/test as 6K/2K2K. In figures \ref{fig:cycle_graph_gaussians} and \ref{fig:noisy_sinusoids} we plot the first 200 timesteps of the training set of "Cycle Graph" and "Correlated Sinusoids" datasets respectively. For "Correlated Sinusoids" we plot the specific case of $N=10$ nodes and two clusters with five nodes each $C=\{5, 5\}$.

\subsection{Implementation details} \label{ap:inference_implementation_details}

In the Inferred Graph Analysis, all models have been trained for 100 epochs, learning rate of $2\cdot10^{-3}$, weight decay $10^{-14}$. The context length (input length of the sequence) is 6, the prediction length (output length of the sequence) is 1. The Mean Absolute Error between prediction and ground truth has been minimized for training. We also added a small regularization value to the loss $R = \frac{10^{-8}}{\# edges} \sum_{i,j} abs(A_{i,j})$ that pushes unnecessary edges closer to 0 for sharper visualizations.

In both BP-GNN and FC-GNN we used the same MLP encoder than the one used METR-LA and PEMS-BAY experiments defined in the Appendix section \ref{ap:architecture_choices}. The Graph Convolutional layer consists of only 1 layer. Since the edge inference is dynamic we obtained the reported Adjacency matrices by averaging  over 10 different $t$.

\newpage
\subsection{Inferred graph for METR-LA} \label{ap:inferred_metrla}

In this section we analyze the adjacency matrices inferred by our FC-GNN method in METR-LA. METR-LA contains 207 nodes. As in the previous synthetic experiment, we build the FC-GNN adjacency matrices from the inferred values $A_{ij}=\alpha_{ij}$. We used the exact same training settings as in the main experiment section \ref{sec:main_results}, but this time we only used one graph layer in the FC-GNN module from which we obtained the $\alpha_{ij}$ values. The provided adjacency has been averaged over 10 different $t$ values.

In METR-LA, sensors are spatially located around a city. Therefore, just for comparison, we also report a matrix built from the distance between each pair of sensors. We call it $A_{sim}$, where each input $A_{sim}[i,j]$ is proportional to the negative distance between the two sensors  $A_{sim}[i,j] = \text{bias} - \text{dist}(i,j)$. We may expect to see some correlations between this matrix and the inferred one, but this does not have to be the case. Close sensors can be correlated, but also far away sensors can be correlated (e.g. we can expect far away residential areas to have a simultaneous increase in traffic right before working hours). Matrices are presented in the following figure, $A_{sim}$ and the inferred adjacency $A$.

\begin{figure}[h!]
\center
\includegraphics[width=0.95\linewidth]{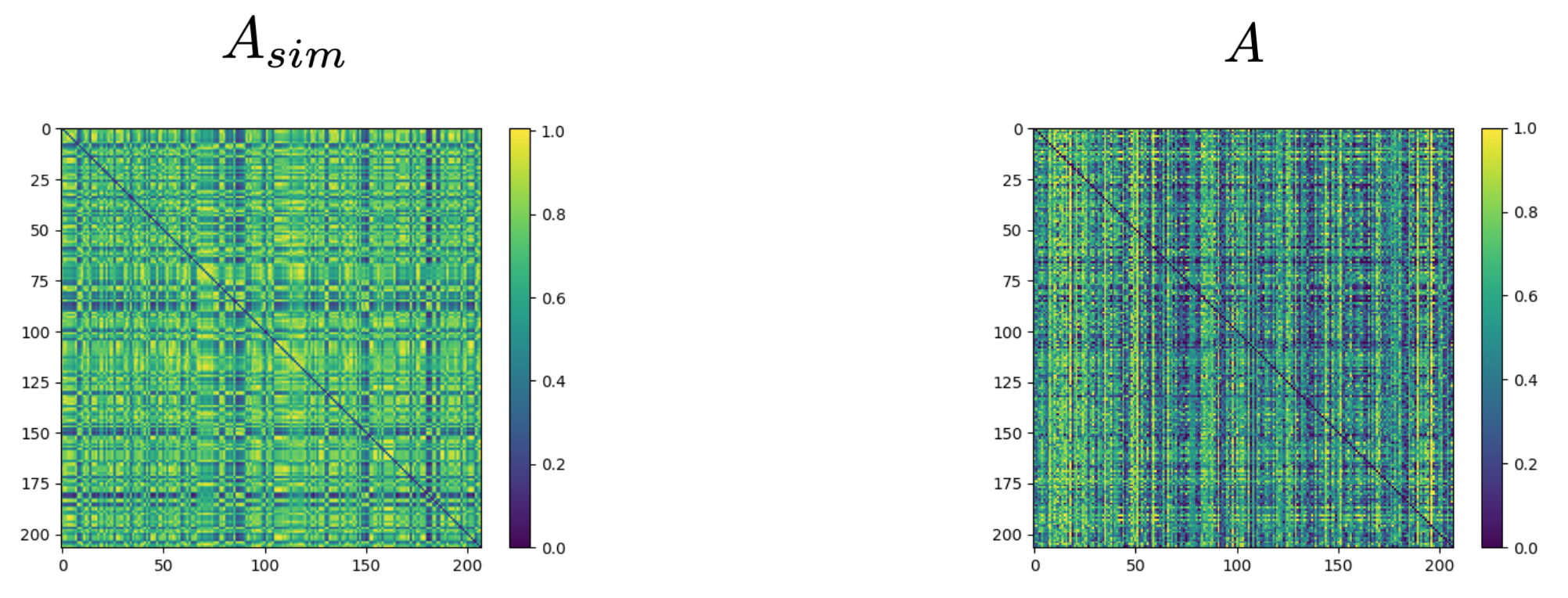}
\caption{$A_{sim}$: matrix obtained from the negative distance among sensors. $A$: Inferred adjacency matrix by FC-GNN.}
\end{figure}

Additionally, our model infers matrices dynamically for each $t$. This means that the inferred adjacency matrix can change depending on the input $\rmx_{t_0:t}$ for different $t$ values. Following, we plot three different matrices for different $t$ values and we see that despite the overall adjacencies share similarities for different $t$, some components differ as we change $t$.

\begin{figure}[h!]
\center
\includegraphics[width=0.95\linewidth]{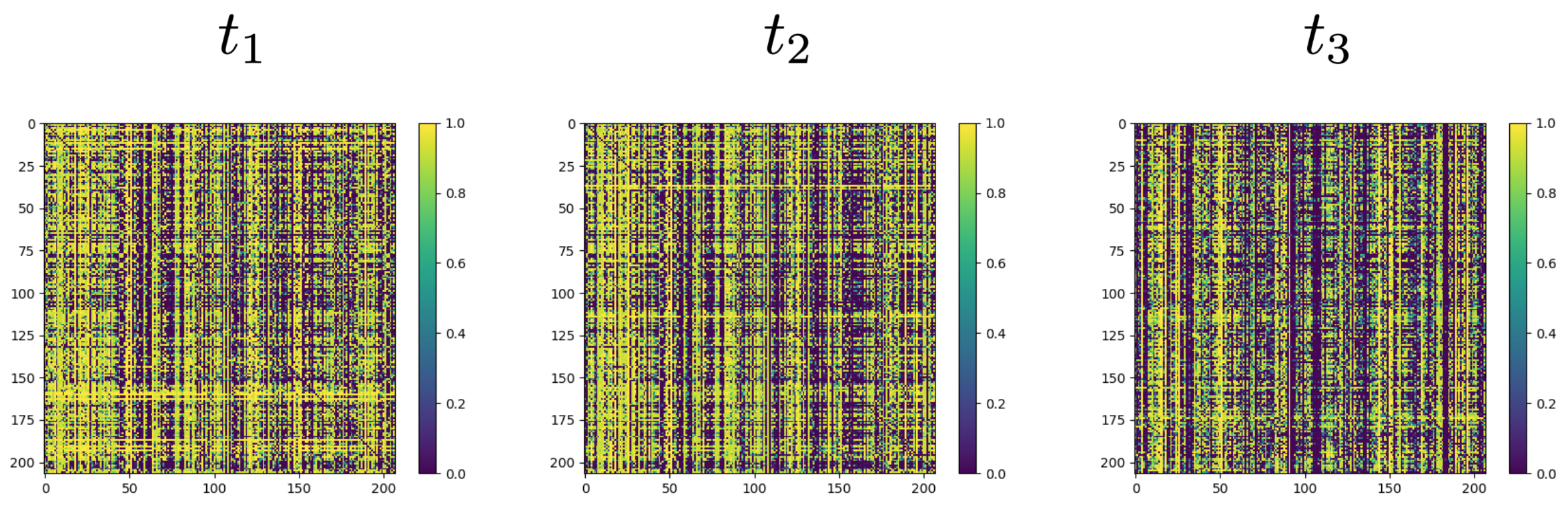}
\caption{Ajacency matrices for different time steps $t$.}
\end{figure}